\documentclass[10pt,twocolumn,letterpaper]{article}

\usepackage[pagenumbers]{cvpr} %

\usepackage{graphicx}
\usepackage{amsmath}
\usepackage{amssymb}
\usepackage{booktabs}
\usepackage{comment}
\usepackage{xcolor}
\usepackage{multirow}
\usepackage{shortbold}
\usepackage{verbatim}
\usepackage{float} %
\usepackage{pifont}
\usepackage[accsupp]{axessibility} %

\DeclareMathOperator*{\argmax}{arg\,max}

\newcommand{\bfpar}[1]{{\vspace{1mm} \par \noindent \bf{{#1}}}}

\definecolor{AccessibleBlue}{rgb}{0.10196, 0.52157, 1.0}
\definecolor{AccessibleRed}{rgb}{0.8314, 0.0667, 0.3490}

\definecolor{msgreenline}{rgb}{0.304,0.367,0.175}
\definecolor{msredline}{rgb}{0.376,0.157,0.151}
\definecolor{msblueline}{rgb}{0.155,0.253,0.371}
\definecolor{msorangeline}{rgb}{0.484,0.294,0.137}
\definecolor{mspurpleline}{rgb}{0.251,0.196,0.318}

\definecolor{msgreen}{rgb}{0.608,0.733,0.349}
\definecolor{msred}{rgb}{0.753,0.314,0.302}
\definecolor{msblue}{rgb}{0.310,0.506,0.741}
\definecolor{msorange}{rgb}{0.969,0.588,0.275}
\definecolor{mspurple}{rgb}{0.502,0.392,0.635}

\definecolor{dgreen}{rgb}{0.,0.596,0.}

\newcommand*\colourcheck[1]{%
  \expandafter\newcommand\csname #1check\endcsname{\textcolor{#1}{\ding{52}}}%
}
\colourcheck{blue}
\colourcheck{green}
\colourcheck{dgreen}
\colourcheck{msgreen}
\colourcheck{msgreenline}
\colourcheck{red}
\newcommand*\colourcross[1]{%
  \expandafter\newcommand\csname #1cross\endcsname{\textcolor{#1}{\ding{55}}}%
}
\colourcross{red}
\colourcross{msred}

\usepackage[pagebackref,breaklinks,colorlinks]{hyperref}

\usepackage[capitalize]{cleveref}
\crefname{section}{Sec.}{Secs.}
\Crefname{section}{Section}{Sections}
\Crefname{table}{Table}{Tables}
\crefname{table}{Tab.}{Tabs.}

\def\cvprPaperID{1100} %
\def\confName{CVPR}
\def\confYear{2022}

\makeatletter
\g@addto@macro\@maketitle{
\vspace{-2.5em}
\begin{figure}[H]
   \setlength{\linewidth}{\textwidth}
\setlength{\hsize}{\textwidth}
\centering
\includegraphics[width=\linewidth]{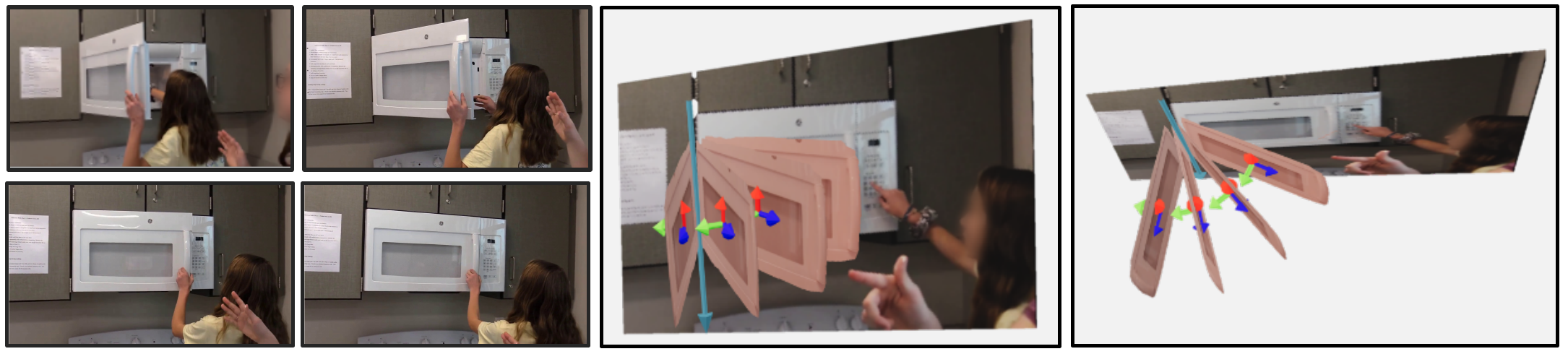}
\caption{Given an ordinary video, our system produces a 3D planar representation of the observed articulation. The 3D renderings illustrate how the microwave (in \textbf{\textcolor{AccessibleRed}{Pink}}) can be articulated in 3D space. We also show the predicted rotation axis using a \textbf{\textcolor{AccessibleBlue}{Blue}} arrow.}
\label{fig:teaser}
\end{figure}
}
\makeatother

\begin{document}

\title{Understanding 3D Object Articulation in Internet Videos}

\author{
Shengyi Qian \kern15pt Linyi Jin \kern15pt Chris Rockwell \kern15pt Siyi Chen \kern15pt David F. Fouhey\\
   University of Michigan\\
	{\tt\small \{syqian,jinlinyi,cnris,siyich,fouhey\}@umich.edu}\\
   {\small \url{https://jasonqsy.github.io/Articulation3D}}
}

\maketitle

\begin{abstract}
We propose to investigate detecting and characterizing the 3D planar articulation of objects from ordinary RGB videos. While seemingly easy for humans, this problem poses many challenges for computers. Our approach is based on a top-down detection system that finds planes that can be articulated. This approach is followed by optimizing for a 3D plane that explains a sequence of detected articulations.
We show that this system can be trained on a combination of videos and 3D scan datasets. When tested on a dataset of challenging Internet videos and the Charades dataset, our approach obtains strong performance.
\end{abstract}

\section{Introduction}
\label{sec:intro}
How would you make sense of Figure~\ref{fig:teaser}? Behind the set of RGB pixels that make up the video is a real 3D transformation consisting of a 3D planar door rotating about an axis. The goal of this paper is to give the same ability to computers. We focus on planar articulation taking the form of a rotation or translation along an axis. This special case of articulation is ubiquitous in human scenes and understanding it lets a system understand objects ranging from refrigerators and drawers to closets and cabinets. While we often learn about these shapes and articulations with physical embodiment~\cite{smith2005development}, we have no difficulty understanding them from video cues alone, for instance while watching a movie or seeing another person perform an action. 
We formalize this ability for computers as recognizing and characterizing a class-agnostic planar articulation via a 3D planar segment, articulation type (rotation or translation), 3D articulation axis, and articulation angle.

This problem is beyond the current state of the art in scene understanding since it requires reconciling single image 3D understanding with {\it dynamic 3D} understanding. While there has been substantial work on 3D reconstruction from a single image~\cite{eigen2015predicting,Wang15,choy20163d,Girdhar16b}, including work dedicated to planes~\cite{liu2019planercnn}, these works focus on reconstructing static scenes. On the other hand, while there has been work understanding articulation, these works often require the placement of tags for tracking~\cite{perez2017c,liu2019learning},
a complete 3D model or depth sensor~\cite{li2020category,mu2021sdf,jain2020screwnet}, or successful 3D human reconstruction~\cite{xu2021d3dhoi}. 
Moreover, making progress is challenging because of data. 
Unsupervised approaches based on motion analysis~\cite{sturm2011probabilistic,pillai2014learning} require something to track, which breaks in realistic data since many human-made articulated objects are untextured (e.g., refrigerators) or transparent (e.g., ovens). 
While supervised approaches~\cite{mu2021sdf,mo2021where2act,li2020category} can perhaps bypass tracking features, they seemingly require access to large amounts of RGBD data of interactions. For now, this data does not exist, and training on synthetic data can fall short when tested on real data (as our experiments empirically demonstrate).

We overcome these challenges with a learning-based approach that combines both detection and 3D optimization and is trained with supervision from multiple sources (Section~\ref{sec:approach}). The foundation of our approach is a top-down detection approach that recognizes articulation axes and types and 3D planes; this approach's outputs are processed with an optimization method that seeks to explain the per-frame results in terms of a single coherent 3D articulation. 

Via this model, we show that one can build an understanding of 3D object dynamics via a mix of 2D supervision on Internet videos of objects undergoing articulation as well as 3D supervision on existing 3D datasets that do not depict articulations. To provide 2D supervision, we introduce (Section~\ref{sec:dataset}) a new set of 9447 Creative Commons Internet videos. These videos depict articulation with a variety of objects as well as negative samples and come with sparse frame annotations of articulation boxes, axes, and surface normals that can be used for training and evaluating planar articulation models.

Our experiments (Section~\ref{sec:experiments}) evaluate how well our approach can recognize and characterize articulation. We evaluate on our new dataset of videos as well as the Charades~\cite{Sigurdsson2016} dataset. We compare with a variety of alternate approaches, including bottom-up signals like optical flow~\cite{teed2020raft} and changes in surface normals~\cite{chen2020oasis}, training on synthetic data~\cite{xiang2020sapien}, as well as systems that analyze human-object interaction~\cite{xu2021d3dhoi}. Our approach outperforms these approaches on our data, often even when the baselines are given access to ground-truth location of articulation.

Our primary contributions include: (1) The new task
of detecting 3D object articulation on unconstrained ordinary RGB videos without requiring RGBD video at training time; (2) A dataset of Internet videos, with sparse frame
annotations of articulation boxes, axes, and surface normals
that can be used for training and evaluating planar articulation models; (3) A top-down detection network and optimization to tackle this problem, which has strong performance on the Internet video dataset and Charades.

\section{Related Work}
\label{sec:related}
Our paper proposes to extract 3D models of articulation from ordinary RGB videos. This problem lies at the intersection of 3D vision, learning from videos, and touches on robotics applications. We note that there are specialized approaches for understanding {\it general} articulation (e.g., non-rigid structure from motion~\cite{torresani2008nonrigid}) as well as for understanding specialized motion models (e.g., for a
full human 3D mesh models~\cite{zhang2019phd} or quadrupeds~\cite{kulkarni2020acsm}) or for understanding more general transformations~\cite{Isola2015,Wang2016}. Our work focuses on understanding the articulation of general objects whose articulated pieces can be represented by a 3D plane rotating or translating.

Due to the ubiquitous nature of articulated objects, the task of understanding them has long been an interest across all of artificial intelligence. In vision, the understanding of the motion of rigid objects undergoing transformations was one of the early successes of computer vision~\cite{tomasi1992shape,kanatani2001motion,wang1994representing}. Unfortunately, these early works rely on reliable motion tracks, which is made difficult by the textureless or reflective nature of many indoor planes (e.g., refrigerator doors). Our top-down detector gives 3D planes that can help provide correspondence between frames where correspondence  is challenging. 

More recent work 
in robotics has used the value of 3D and integrated it into their modeling approaches~\cite{sturm2011probabilistic,pillai2014learning,cifuentes2016probabilistic,desingh2019factored,michel2015pose}; 
however their approaches often use an RGBD sensor, unlike our use of ordinary RGB sensors. 
This dependence on RGBD has been carried forward to the most recent work that uses deep learning frameworks~\cite{xiang2020sapien,batra2020rearrangement,li2020category,liu2021towards,jain2020screwnet,Wang_2019_CVPR}. 
In fact, some methods require full 3D models \cite{mu2021sdf}, which is typically unavailable in real world 3D scans.
\cite{mo2021where2act} by Mo et al. can be run on 2D images as long as the point cloud encoder is replaced with a RGB encoder, but its 2D images contain a single object without any background, instead of challenging Internet videos. While there has been increasing amounts of work aimed at virtual articulated objects~\cite{szot2021habitat,xiang2020sapien}, simultaneously achieving scale and quality is challenging. For instance ReplicaCAD~\cite{szot2021habitat} has only 92 objects.
In contrast, our approach works at test time on standard RGB videos by bringing its own 3D via a learned detector~\cite{liu2019planercnn} trained on RGBD data~\cite{dai2017scannet}.

While our outputs are 3D planar regions, our approach is deeply connected to the task of understanding human-object interactions. In these works~\cite{Chao15,Gkioxari18,Shan20}, the goal is to recognize the relationship between humans and the objects they interact with. The interactions that we study are caused by these humans, and so we use an approach that can predict human-object interactions~\cite{Shan20} to help identify the data we train our systems on. The most related work in this area is~\cite{xu2021d3dhoi}, which aims to jointly understand dynamic 3D human-object interactions in 3D.
This work, however assumes that the object CAD model is known once the articulated object is detected, which we do not need. Our method also works with articulation videos that are more varied in viewpoint and perspective.
A more thorough understanding of the joint relationship between articulated objects and human-object interaction, akin to early work~\cite{Koppula2013,Fouhey12}, is beyond the scope of this work, but of future value.

We solve the problem of describing 3D articulation by producing 3D planar models. This uses advances in 3D from a single image. 
In particular, we build on PlaneRCNN~\cite{liu2019planercnn}, which is part of a growing body of works aimed at extracting planes from single images~\cite{yang2018recovering,YuZLZG19,liu2018planenet}. These planes have advantages for the articulation reasoning since they offer a compact representation to track and describe. While we use plane recognition, the plane is just one component of our output (along with rotation axes) and we analyze our output in a video with temporal optimization.

\section{Dataset} %
\label{sec:dataset}
\begin{figure*}[t]
    \centering
    \includegraphics[width=\linewidth]{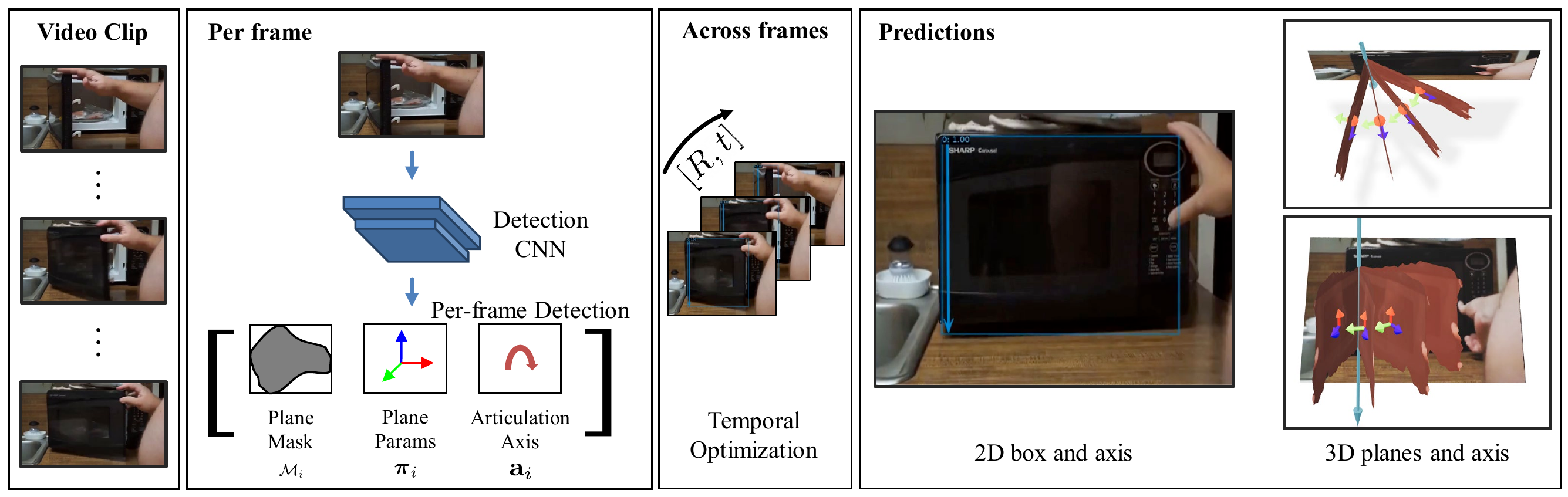}
    \caption{Overview of our approach. (a) Given an ordinary video clip, we first apply our 3D Articulation Detection Network (3DADN) to detect 3D planes can be articulated for each frame. (b) We then apply temporal optimization to fit the articulation model. Final results are demonstrated in both 2D image and 3D rendering.}
    \label{fig:approach}
 \end{figure*}

One critical component of our approach is accurate 2D annotations of articulation occurring in RGB data. We show that these 2D annotations can be combined with existing RGBD data and the right method to build systems that understand 3D articulation on video data. We next describe how we collect a dataset of articulations.
Our goals are to have a large collection of annotations of object box, articulation type, and axis. 
Rather than directly look for 
examples of people articulating objects, we follow the data-first approach of~\cite{Fouhey18,zellers2019recognition,Shan20,Damen18}, namely to
gather data containing many related activities and then analyze and annotate it.

\bfpar{Data Collection.}
Our pipeline generates a set of candidate clips to be annotated from a collection of candidate videos via an automatic pipeline that aims to eliminate frames that are easy to see as not depicting articulation. We begin with candidate videos that are found by variants of searches for a set of 10 objects among Creative Commons videos on YouTube. Within these videos, we find stationary continuous shots in these videos with homographies~\cite{Hartley04} fit on ORB \cite{rublee2011orb} features. Many of these clips cannot depict interaction since they do not contain any people or do not contain the objects of interest. We filter by responses by a hand detector~\cite{Shan20} trained on $100$K+ frames of Internet data, as well as object detectors trained on COCO~\cite{Lin2014} and LVIS~\cite{gupta2019lvis}. These filtering steps maximize the use of annotator time by eliminating clear negatives, and generate a large number of candidate clips. 

With a collection of candidate clips of interest, we then turn to manual annotation. For a given clip, we hire an annotation company to annotate frames sparsely (every 10 frames) within the clip. They annotate: ({\bf box}) a box around the articulated plane and its type, if it exists; and ({\bf axis}) the projection of the articulation axis, framed as a line segmentation annotation problem. This results in a set of 19411 frames, containing 19411 boxes around articulating planes with 13508 rotation axes and 2755 translation axes, as well as 39411 negative frames. 
The number of articulation axes is not equal to the number of boxes, since some articulation axes are outside the image.
We provide training, validation, and test splits based on uploader, leading to 7845/601/1001 videos in the train/val/test split. A more complete description of our annotation pipeline appears in the supplement.

We collect two additional annotations. 
For the test set, we also annotate the surface normal of the plane following~\cite{chen2020oasis}, so we can evaluate how well our model can learn 3D properties. 
To show generalization, we also collect the same annotations except surface normals on the Charades~\cite{Sigurdsson2016} dataset.

\bfpar{Data Availability and Ethics.}
Our data consists of videos that users uploaded publicly and chose to share as Creative Commons data. These do not involve interaction with humans or private data. We filtered obviously offensive content, videos depicting children, and cartoons. Examples appear throughout the paper; screenshots of annotation instructions and details appear in the supplement.

\section{Approach}
\label{sec:approach}

The goal of our approach is to detect and characterize planar articulation in an unseen RGB video clip. These articulations are an important special case that are ubiquitous in human scenes. As shown in Figure~\ref{fig:approach}, we propose a {\it 3D Articulation Detection Network} (3DADN) to solve the task. As output, the 3DADN produces the type of motion (rotation or translation), a bounding box around where the motion is located, the 2D location of the rotation or translation axis, and the 3D location of the articulated plane. The 3DADN's output is followed by post-processing to find a consistent explanation over the whole video.

\subsection{3D Articulation Detection Network}
\label{sec:approach_detect}

The 3DADN processes each frame independently. Its output consists of: a segment mask $\mathcal{M}_i$; plane parameters $\piB_i = [\nB_i, o_i]$ giving the plane equation $\piB_i^T [x,y,z,-1] = 0$ (where $\nB_i$ is the the plane normal with $||\nB_i||_2=1$ and $o_i$ is the plane offset); a projected rotation or translation axis $\aB_i = [\theta, p]$ which is the projection of the 3D articulation axis; and articulation type.

We use a top-down approach to detect this representation, which we train on RGB videos that depict articulation {\it without} 3D information as well as RGBD images {\it with} 3D information that do not depict articulation. 
Our backbone is a Faster R-CNN~\cite{ren2015faster} style network that first detects bounding boxes for the articulating objects and classifies them into two classes (rotation and translation).
These boxes provide ROI-pooled features that are passed into detection heads that predict our outputs ($\mathcal{M}_i, \piB_i, \aB_i$).
Our heads and losses for $\mathcal{M}_i$ follow the common practice of Mask R-CNN \cite{He17}.
We describe $\aB_i$ and $\piB_i$ below.

\bfpar{Parameterizing Rotation and Translation Axis.}
We model the projected articulation axis as a 2D line in the image. This projected axis is the projection of the 3D articulation axis (e.g., the hinge of a door). We describe the projected axis with the normal form of the line, 
$x \cos(\theta) + y \sin(\theta) = p$
where $p \ge 0$ is the distance from the box to the center and $\theta$ is the inclination of the normal of the axis in pixel coordinates. Since a translation corresponds to a direction/family of lines as opposed to a line,  we define $p=0$ for translation arbitrarily. 

The articulation head contains two independent branches for predicting the rotation and translation axes. We handle the circularity of the prediction of $\theta$ by lifting predictions and ground-truth for the angle to the 2D unit circle; since the line is 180-degree-ambiguous (i.e., $\theta+\pi$ is the same as $\theta$), we predict a 2D vector $[\sin(2\theta),\cos(2\theta)]$. The resulting network thus predicts a 3D vector containing $\theta$ and $p$, which we supervise with a L1 loss.

\bfpar{Parameterizing Plane Parameters.}
Following a line of work on predicting planes in images, we use a 3D plane \cite{liu2018planenet} to represent the 3D locations of the articulated objects, since many common articulated objects like doors, refrigerators, and microwaves can be modeled as planes and because past literature~\cite{liu2019planercnn,jiang2020peek,jin2021planar} suggests that R-CNN-style networks are adept at predicting plane representations.

A 3D plane is represented by plane parameters $\piB_i = [\nB_i, o_i]$ giving the plane equation $\piB_i^T [x,y,z,-1] = 0$ 
With camera intrinsics, planes can be recovered in 3D and with a mask, this plane can be converted to a plane segment.
Following \cite{liu2019planercnn,jin2021planar}, we extend R-CNN by adding a plane head which directly regresses the normal of the plane. A depth head is used to predict depth of the image. The depth is only used to calculate the offset value of the plane. We supervise with L2 loss for the plane normal regression and L1 loss for the depth regression.

\bfpar{Training.} 
There is no dataset that is non-synthetic and large enough to train a 3DADN directly: the 3DADN needs both realistic interactions and 3D information. 
However, we can train the 3DADN in parts. 
In the first stage, we train the backbone, RPN, and axis heads directly on our Internet video training set, which has boxes and axes. We then freeze the backbone, RPN, and axis heads and fine-tune the mask and plane head on a modified version of ScanNet~\cite{dai2017scannet}.

In particular, we found that humans often occlude the objects they articulate and models that had not seen humans in training produced worse qualitative results. 
We therefore composited humans from SURREAL~\cite{varol17_surreal} into the scenes.
We randomly sample 98,235 ScanNet images, select a synthetic human and render it on ScanNet backgrounds.
In training, we do not change the ground truth, pretending the ground truth plane is partially occluded by humans and training our model to identify them.

Meanwhile, we found that the order of training the heads was crucial. Planes in ScanNet~\cite{dai2017scannet} are defined geometrically, and so unopened doors often merge with walls; similarly, ScanNet~\cite{dai2017scannet} does not contain transitional moments during which planes are articulating. Thus, RPNs trained on ScanNet~\cite{dai2017scannet} perform poorly on articulation videos.
Instead, it is important to train the RPN on our Internet videos, freeze the backbone, and only rely on ScanNet to train plane parameters and masks, which are unavailable in Internet videos.
During inference we keep the ScanNet camera since our data does not have camera intrinsics.

\bfpar{Implementation Details.}
Full architectural details of our approach are in the supplemental.
Our model is implemented using Detectron2~\cite{wu2019detectron2}. The backbone uses ResNet50-FPN~\cite{Lin_2017_CVPR} pretrained on COCO~\cite{Lin2014}.

\begin{figure*}
    \centering
    \begin{tabular}{c@{\hskip4pt}c@{\hskip4pt}c@{\hskip4pt}c@{\hskip4pt}c@{\hskip4pt}c}
    \toprule
    Input & Pred1 & Pred2 & Pred3 & Pred4 & Pred 3D \\
    \midrule
    \frame{\includegraphics[width=0.15\linewidth]{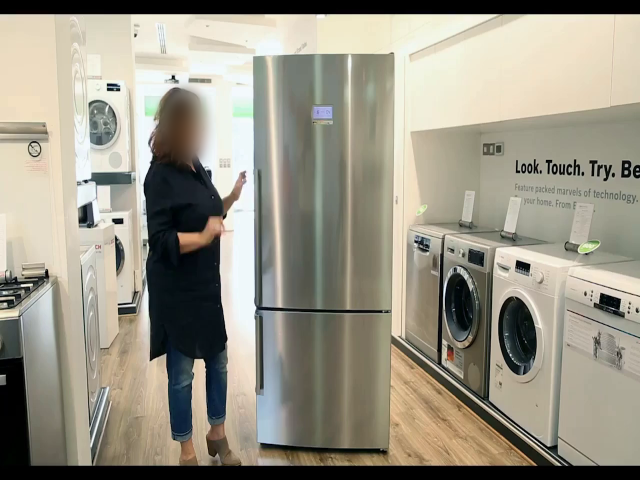}}
    & \frame{\includegraphics[width=0.15\linewidth]{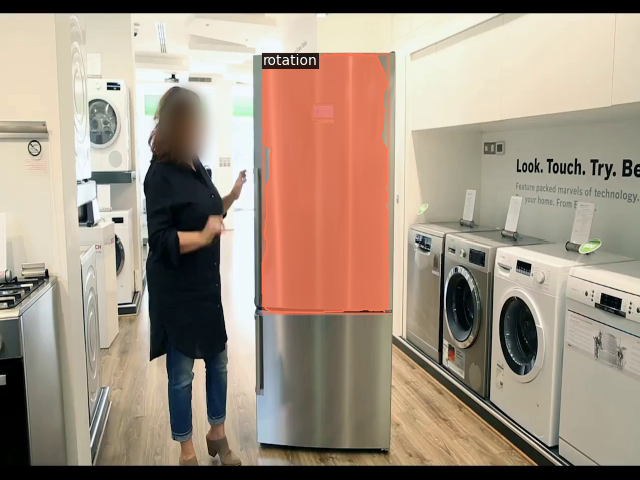}}
    & \frame{\includegraphics[width=0.15\linewidth]{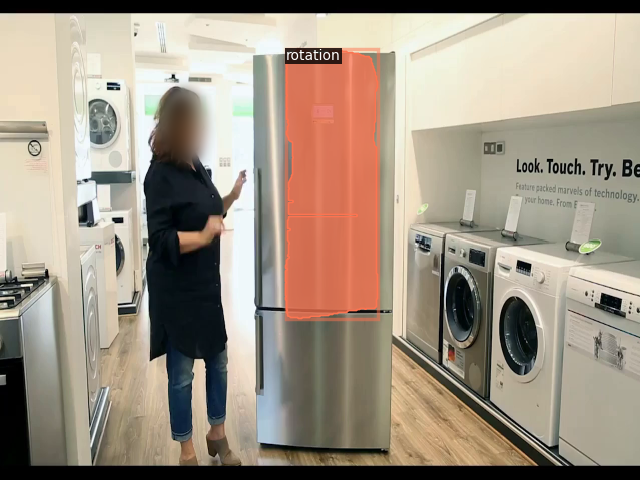}}
    & \frame{\includegraphics[width=0.15\linewidth]{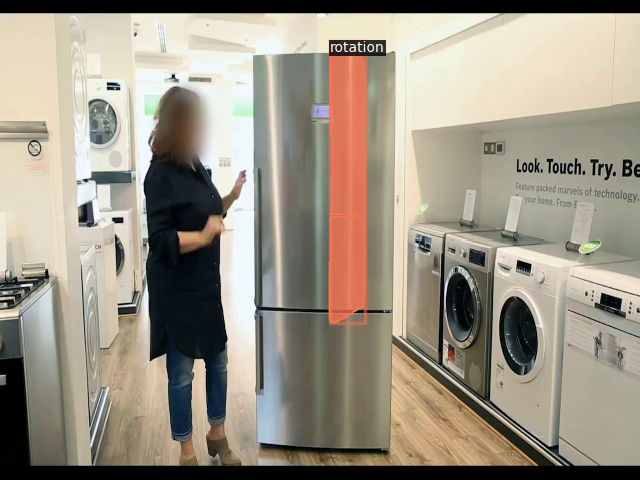}}
    & \frame{\includegraphics[width=0.15\linewidth]{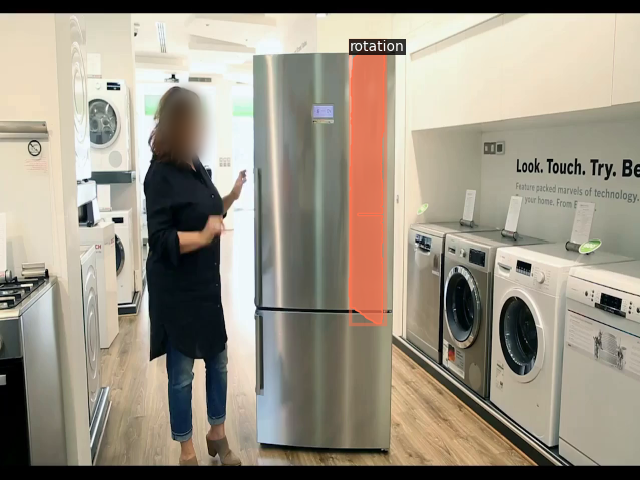}}
    & \frame{\includegraphics[width=0.15\linewidth]{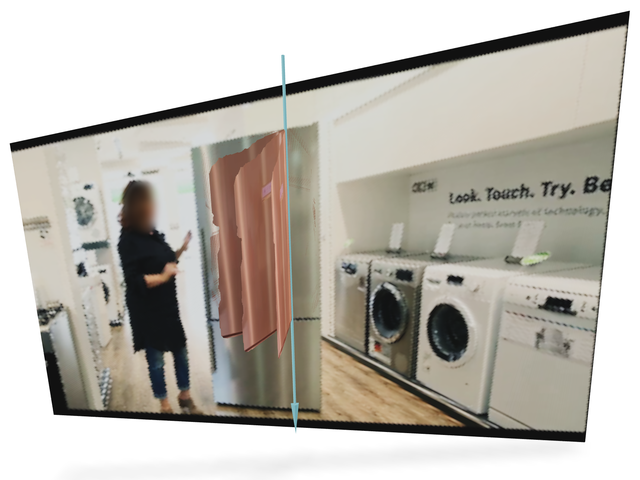}}\\
    
    \frame{\includegraphics[width=0.15\linewidth]{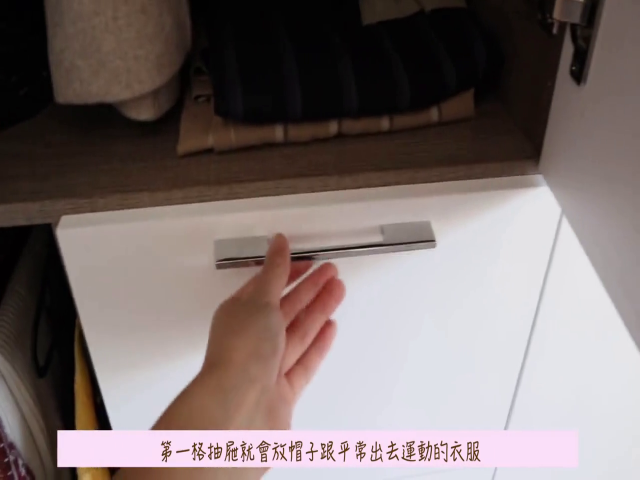}}
    & \frame{\includegraphics[width=0.15\linewidth]{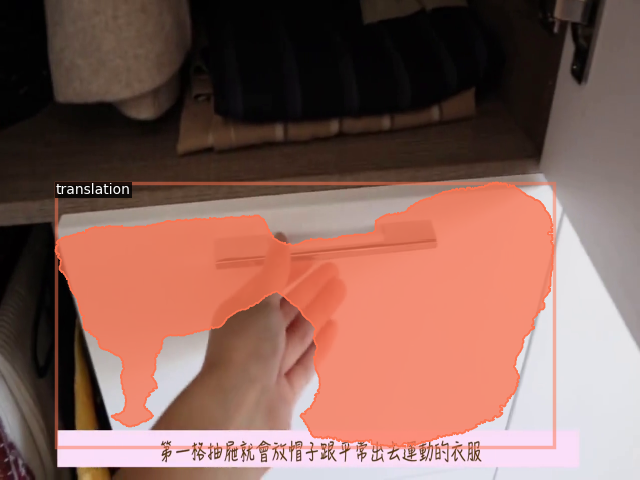}}
    & \frame{\includegraphics[width=0.15\linewidth]{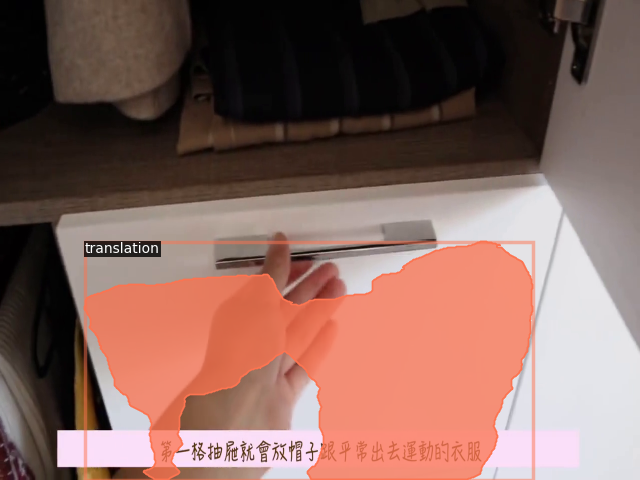}}
    & \frame{\includegraphics[width=0.15\linewidth]{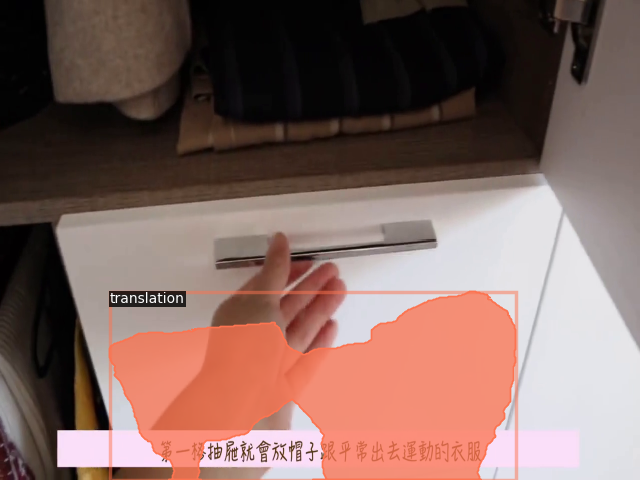}}
    & \frame{\includegraphics[width=0.15\linewidth]{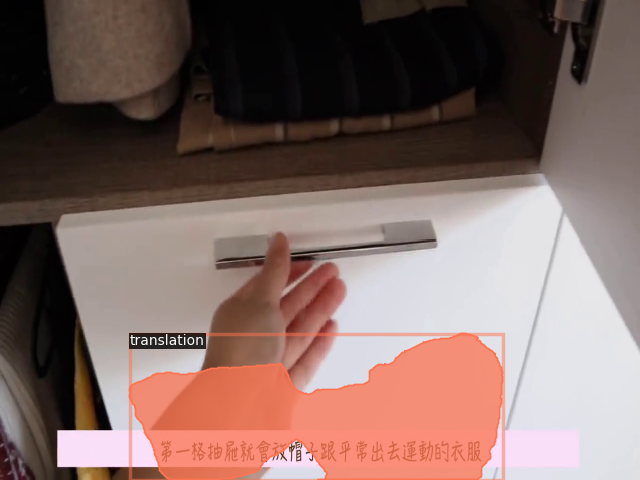}}
    & \frame{\includegraphics[width=0.15\linewidth]{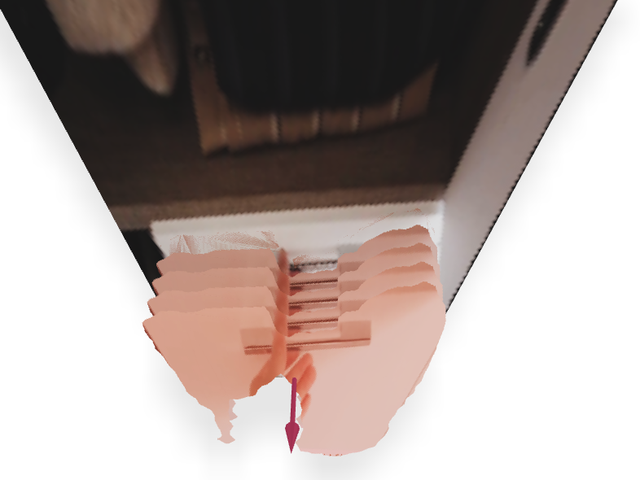}}\\
    \frame{\includegraphics[width=0.15\linewidth]{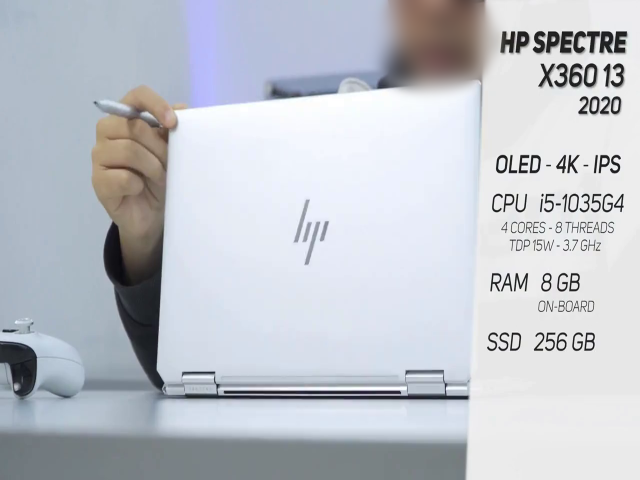}}
    & \frame{\includegraphics[width=0.15\linewidth]{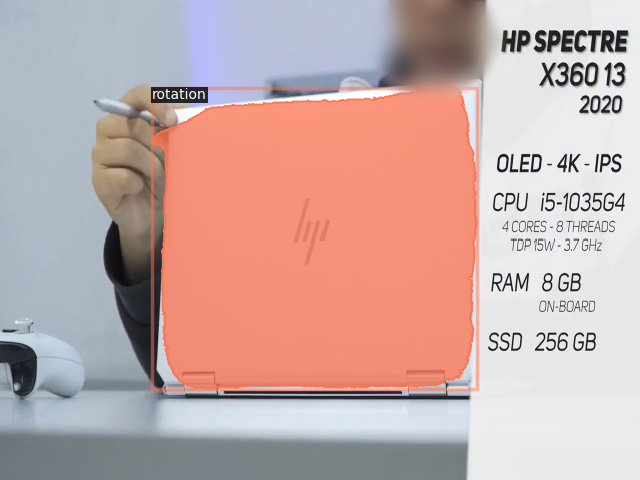}}
    & \frame{\includegraphics[width=0.15\linewidth]{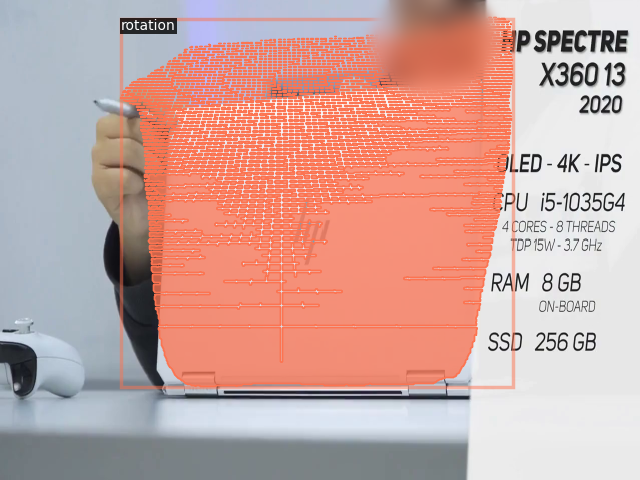}}
    & \frame{\includegraphics[width=0.15\linewidth]{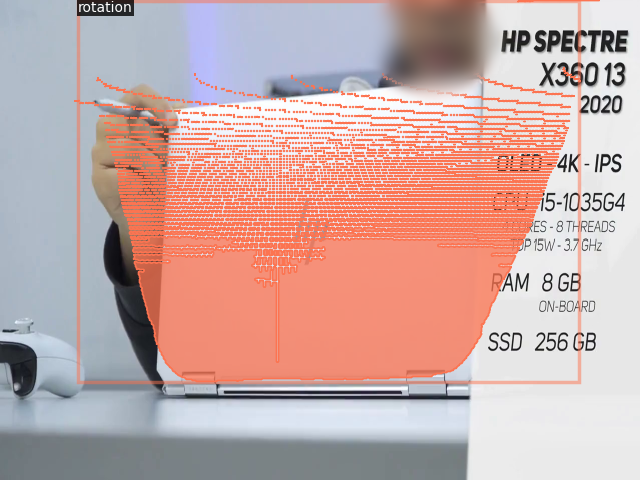}}
    & \frame{\includegraphics[width=0.15\linewidth]{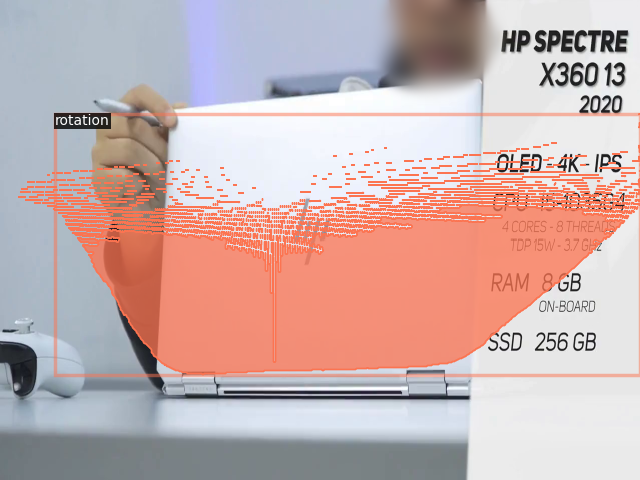}}
    & \frame{\includegraphics[width=0.15\linewidth]{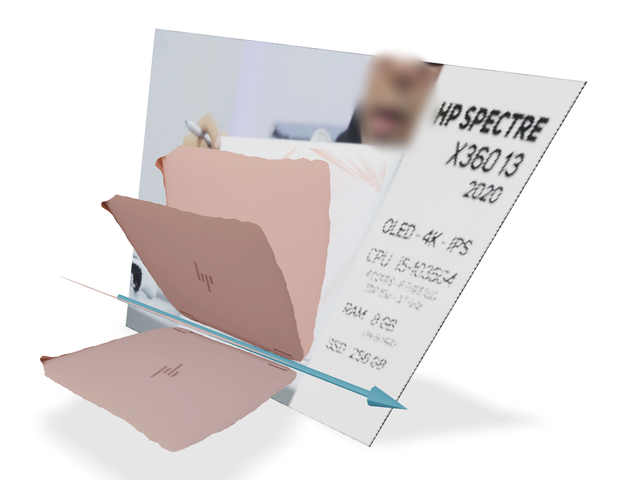}}\\
    \frame{\includegraphics[width=0.15\linewidth]{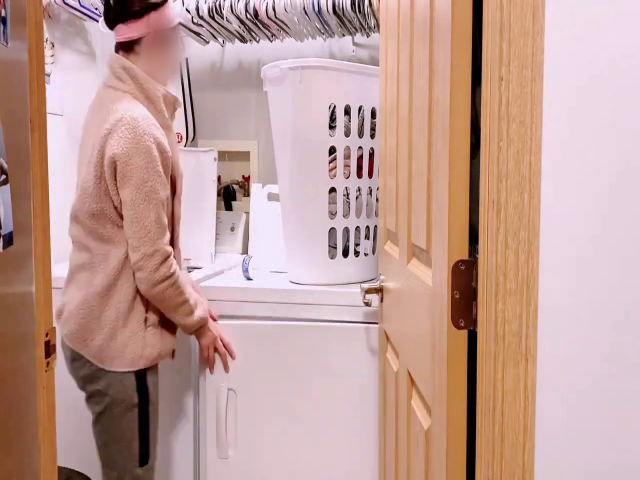}}
    & \frame{\includegraphics[width=0.15\linewidth]{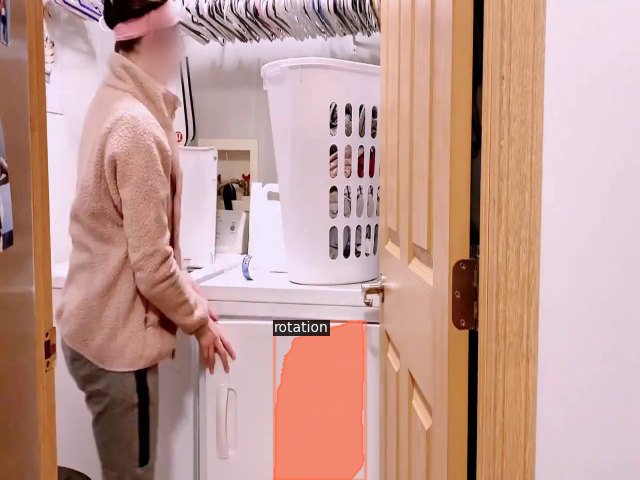}}
    & \frame{\includegraphics[width=0.15\linewidth]{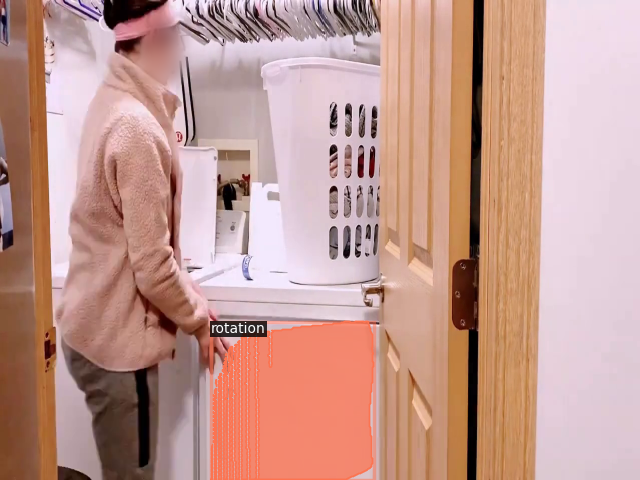}}
    & \frame{\includegraphics[width=0.15\linewidth]{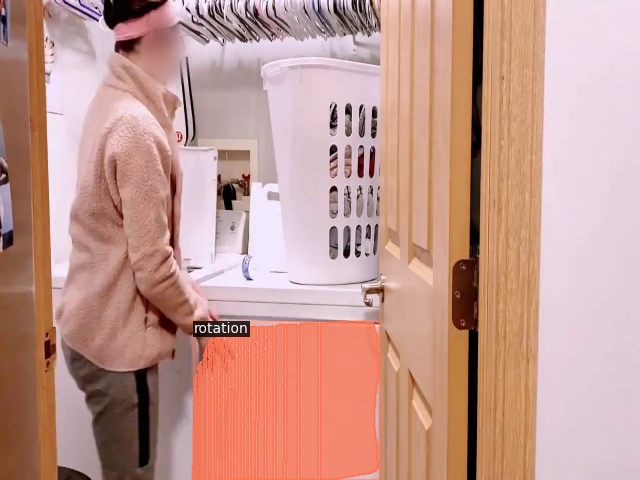}}
    & \frame{\includegraphics[width=0.15\linewidth]{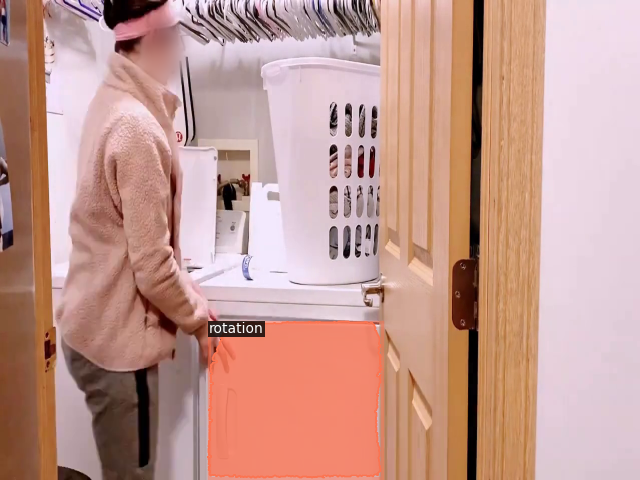}}
    & \frame{\includegraphics[width=0.15\linewidth]{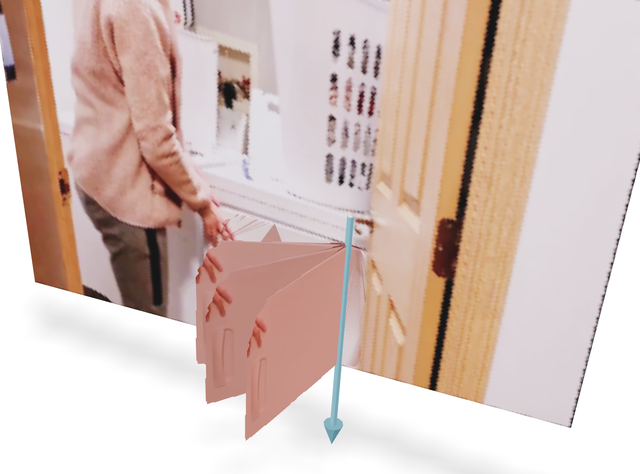}}\\
    \frame{\includegraphics[width=0.15\linewidth]{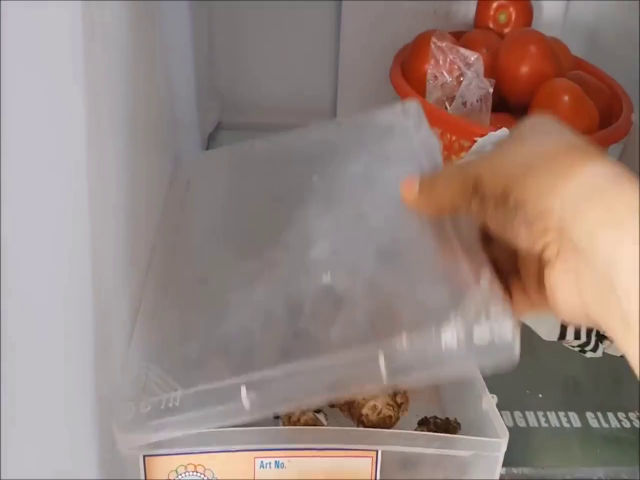}}
    & \frame{\includegraphics[width=0.15\linewidth]{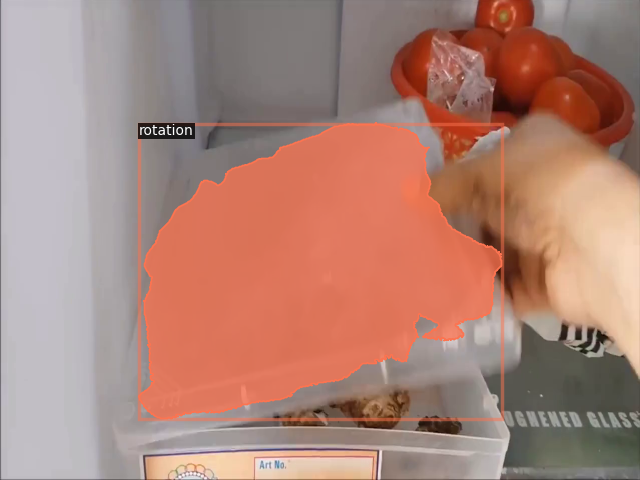}}
    & \frame{\includegraphics[width=0.15\linewidth]{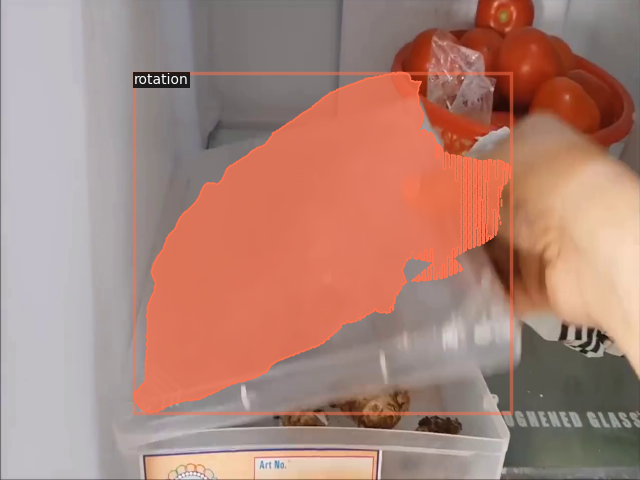}}
    & \frame{\includegraphics[width=0.15\linewidth]{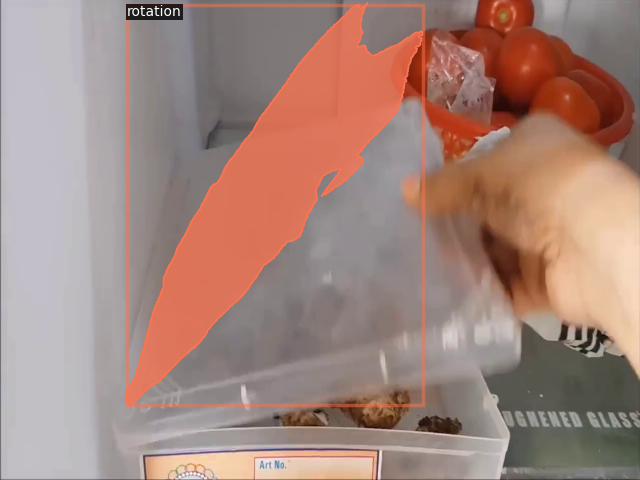}}
    & \frame{\includegraphics[width=0.15\linewidth]{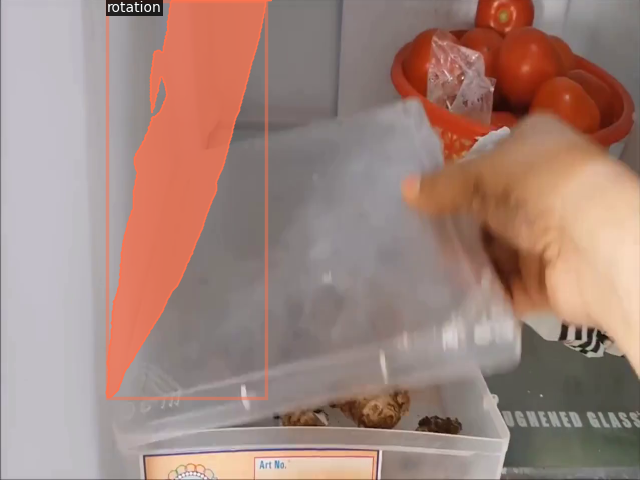}}
    & \frame{\includegraphics[width=0.15\linewidth]{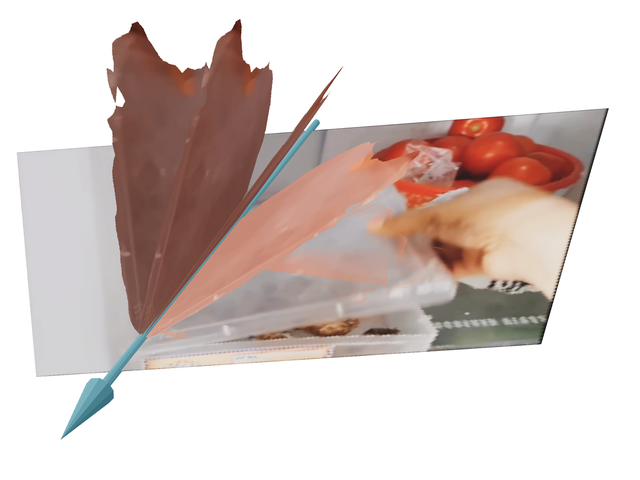}}\\
    \bottomrule
    \end{tabular}
    \caption{Predictions on Internet videos. For each example, we show the input (left), detected 2D planes and how they will be articulated using the predicted articulation axes and surface normals (middle). We also show 3D renderings to illustrate how these common objects are articulated in the 3D space (right). The predicted rotation axis is shown as the \textbf{\textcolor{AccessibleBlue}{Blue}} arrow, and translation axis is the \textbf{\textcolor{AccessibleRed}{Pink}} arrow.
    }
    \label{fig:vis_wall}
\end{figure*}

\subsection{Temporal Optimization}
\label{sec:approach_optimize}

After the 3DADN provides per-frame estimates of articulations, we perform temporal optimization to find a single explanation for the detections across frames. We are given a sequence of detections indexed by a time of the form $[\mathcal{M}_i^{(t)}, \piB_i^{(t)}, \aB_i^{(t)}]$. We aim to find a single consistent explanation for these detections.

\bfpar{Tracking.}
Optimizing requires a sequence of planes to optimize over. We match box $i$ with the box in the next frame according to pairwise intersection over union (IoU). Box $i$ at $t$ matches box $j = \argmax_{j'} \textrm{IoU}(\mathcal{M}^{(t)}_i, \mathcal{M}^{(t+1)}_{j'})$ at time $t+1$; we then track greedily to get a sequence. We subsequently drop the subscripts for clarity.

\bfpar{Articulation Model Fitting.}
Given a sequence of detections, we find a consistent explanation via a RANSAC-like approach. We begin with a hypothesis of a plane segment $\piB$ and articulation axis $\aB$, which we obtain by selecting output on a reference frame. Along with an assumed camera intrinsics $\KB$, the plane parameters let us lift the plane segment and axis to 3D, producing 3D plane segment $\PiB$ and 3D axis $\AB$. Then, for each frame $t$, we solve for an articulation degree
$\alpha^{(t)}$ maximizing the reprojection agreement with the predicted mask at time $t$. Let us define the reprojection score as  
\begin{equation}
r(\alpha,t) = \textrm{IoU}\left(\mathcal{M}^{(t)}, \KB \left[ \RB_\alpha, \tB_\alpha\right] \PiB \right),
\end{equation}
where $\RB_{\AB,\alpha}$ and $\tB_{\AB,\alpha}$ are $\alpha$ steps over the rotation and translation for axis $\AB$. We then solve for $\alpha^{(t)}$ by solving $\argmax_\alpha r(\alpha)$, which gives a per-frame angle using grid search.
We detect articulation by calculating how well the rotation degree $\alpha^{(t)}$ can be explained as a linear function of $t$ (i.e., that there is constant motion). Since many scenes are not constant motion, we have loose thresholds: we consider $R^2 \ge 0.4$ and slope $k>0.1$ to be an articulation. We exclude hypotheses where all $r(\alpha^{(t)},t) < 0.5$.

\section{Experiments}
\label{sec:experiments}

\begin{figure*}
    \centering
    \begin{tabular}{c@{\hskip4pt}c@{\hskip4pt}c@{\hskip4pt}c@{\hskip4pt}c@{\hskip4pt}c}
    \toprule
    Flow+Normal & SAPIEN & SAPIEN w/ gtbox&  D3D-HOI & Ours & GT \\
    \midrule
    \frame{\includegraphics[width=0.15\linewidth]{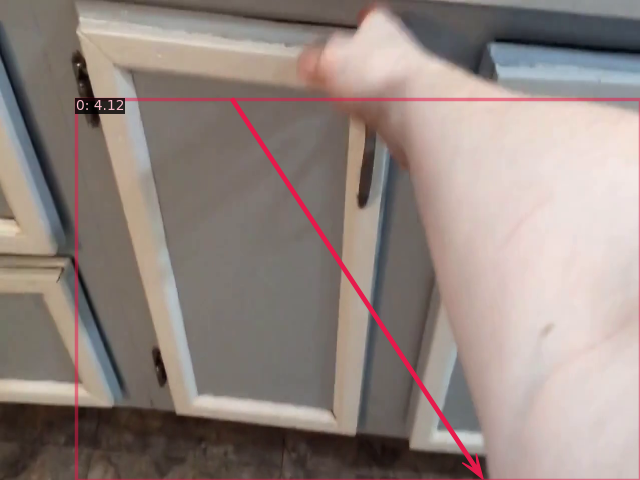}}
    & \frame{\includegraphics[width=0.15\linewidth]{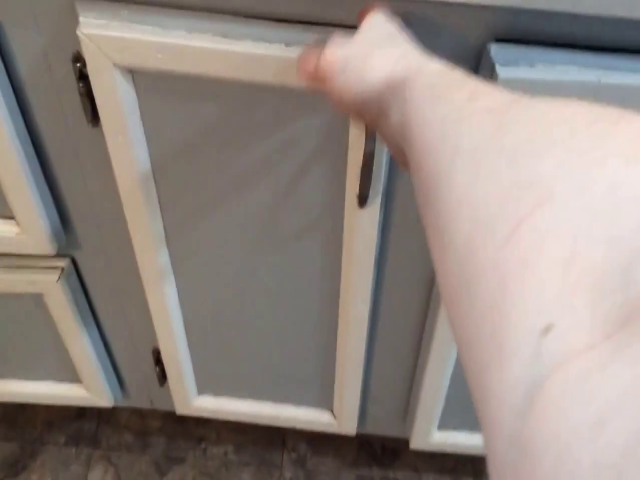}}
    
    & \frame{\includegraphics[width=0.15\linewidth]{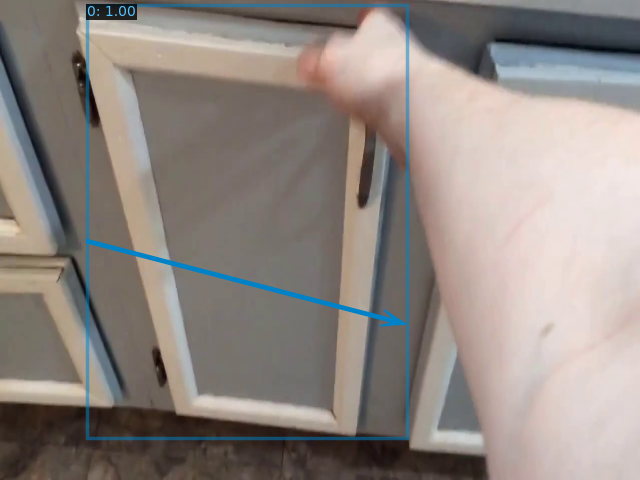}}
    & \frame{\includegraphics[width=0.15\linewidth]{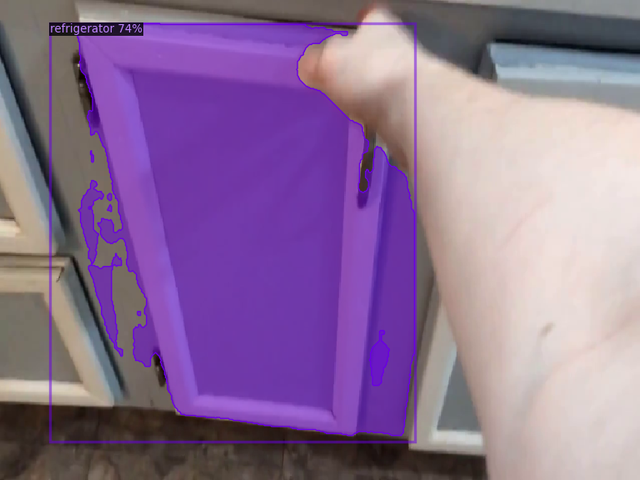}}
    
    & \frame{\includegraphics[width=0.15\linewidth]{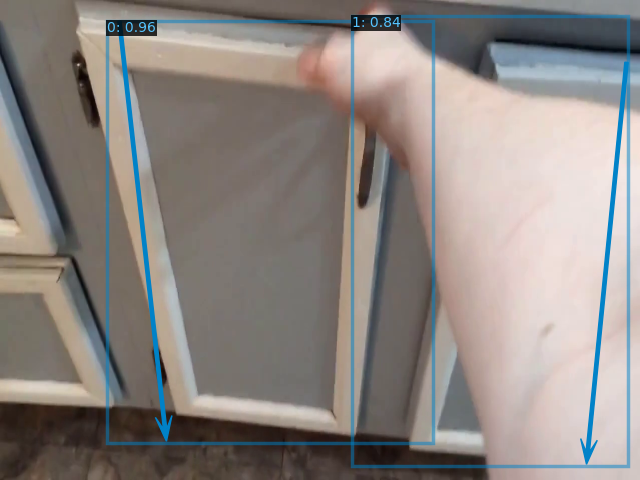}}
    & \frame{\includegraphics[width=0.15\linewidth]{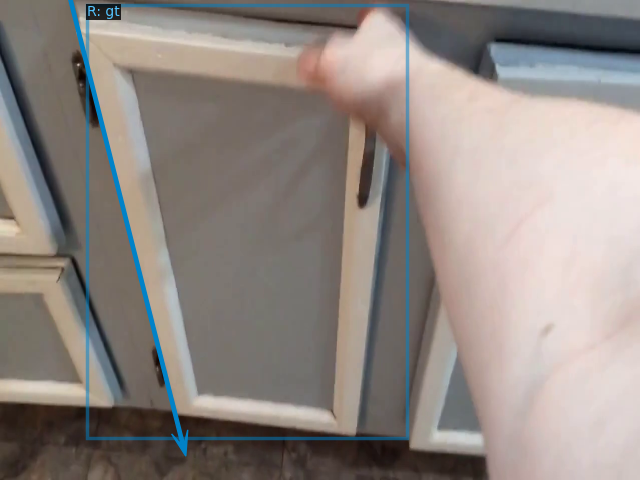}}\\
    
    \frame{\includegraphics[width=0.15\linewidth]{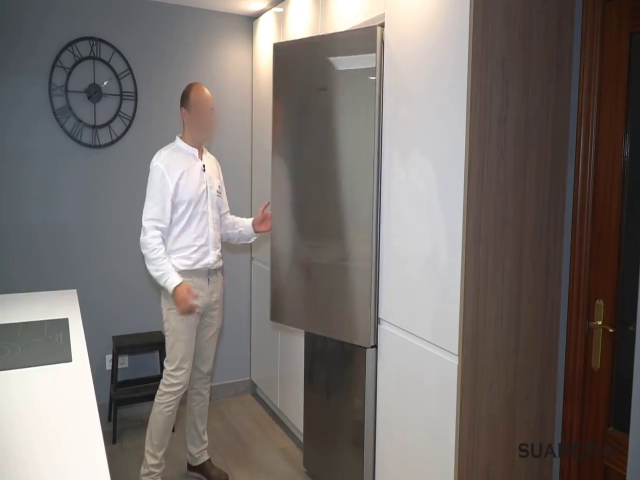}}
    & \frame{\includegraphics[width=0.15\linewidth]{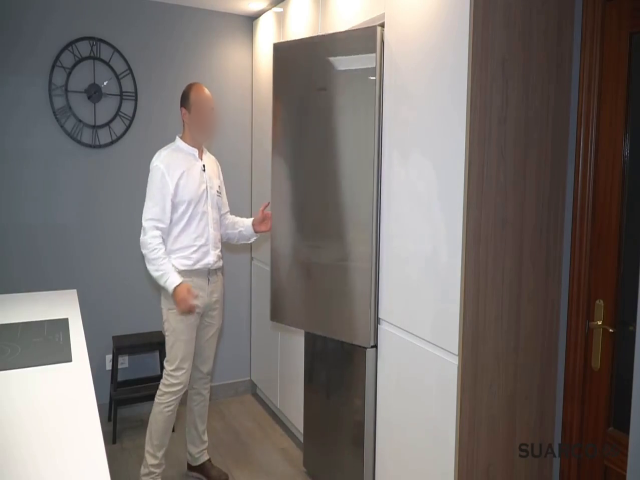}}
    
    & \frame{\includegraphics[width=0.15\linewidth]{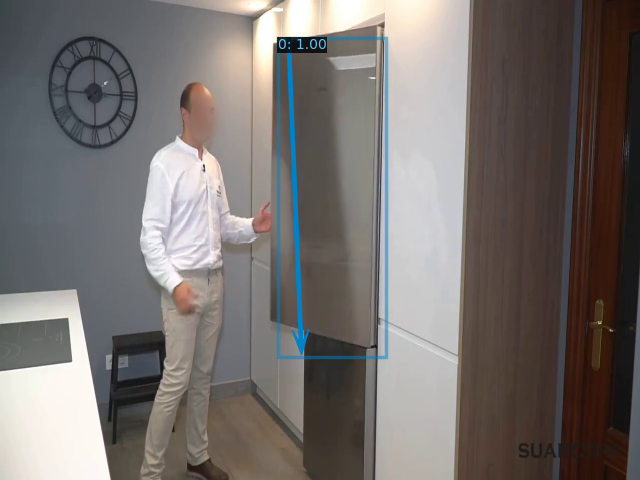}}
    & \frame{\includegraphics[width=0.15\linewidth]{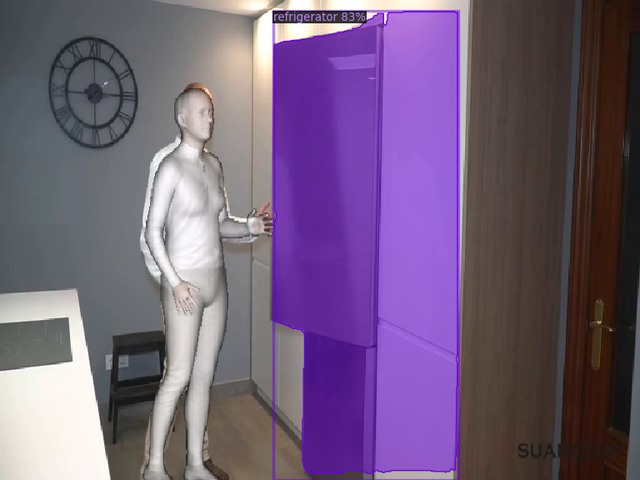}}
    
    & \frame{\includegraphics[width=0.15\linewidth]{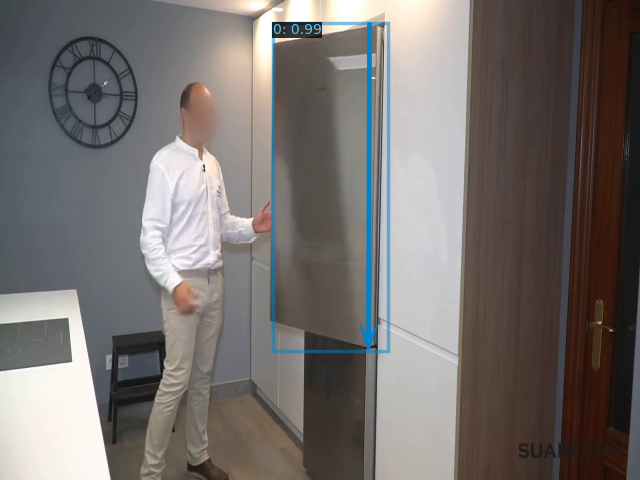}}
    & \frame{\includegraphics[width=0.15\linewidth]{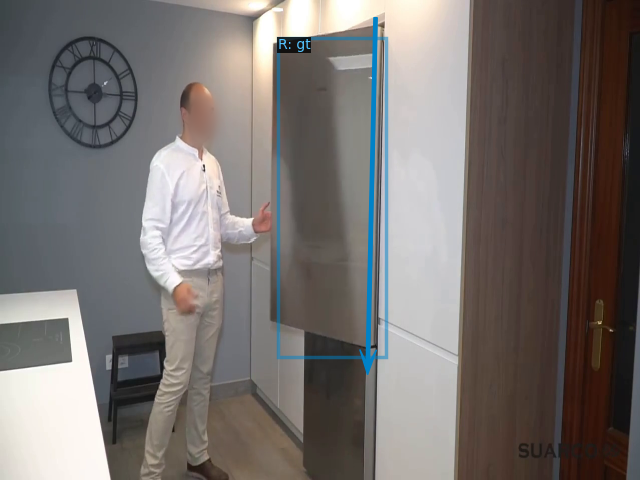}}\\
    
    \frame{\includegraphics[width=0.15\linewidth]{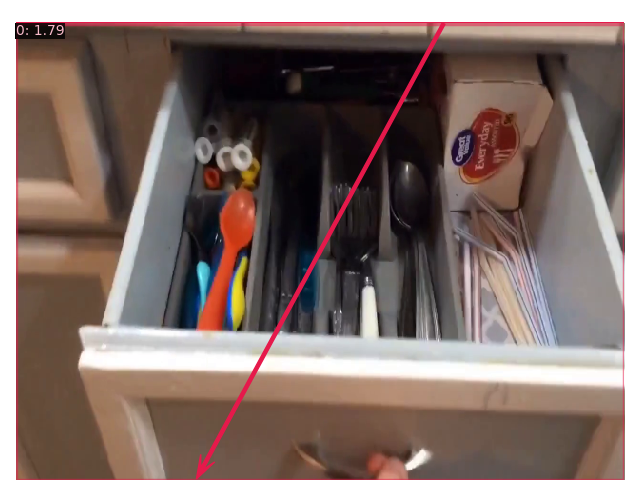}}
    & \frame{\includegraphics[width=0.15\linewidth]{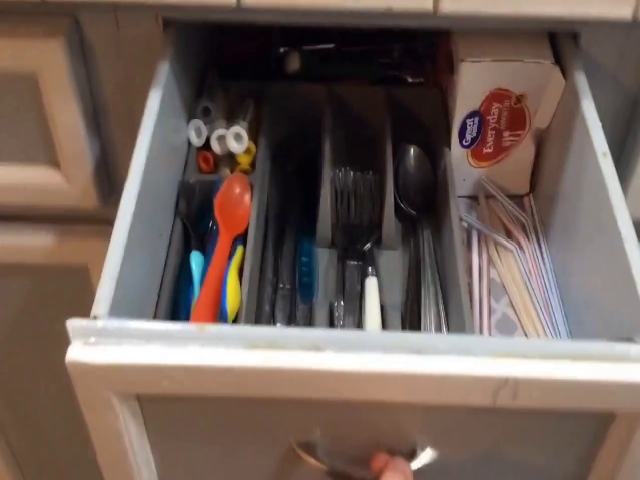}}
    
    & \frame{\includegraphics[width=0.15\linewidth]{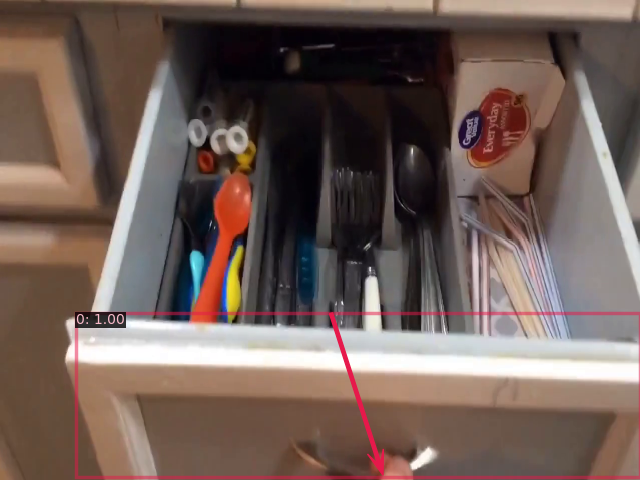}}
    & \frame{\includegraphics[width=0.15\linewidth]{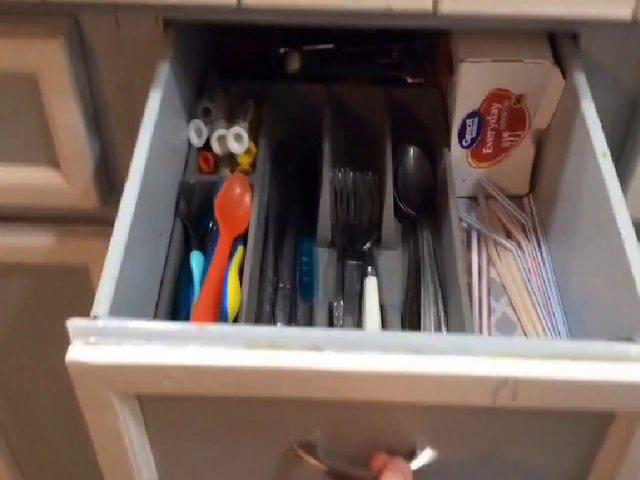}}
    
    & \frame{\includegraphics[width=0.15\linewidth]{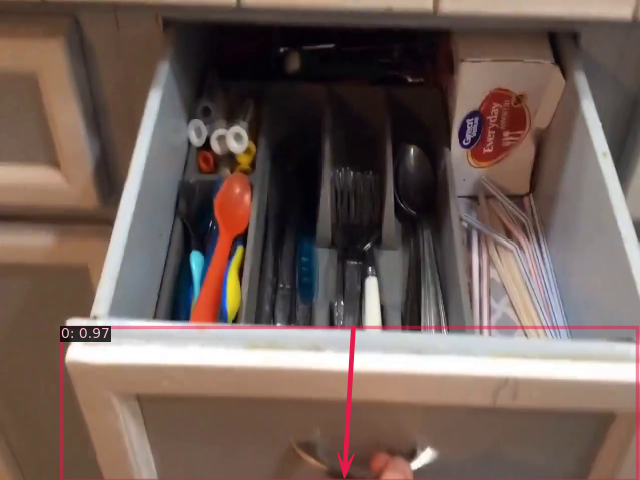}}
    & \frame{\includegraphics[width=0.15\linewidth]{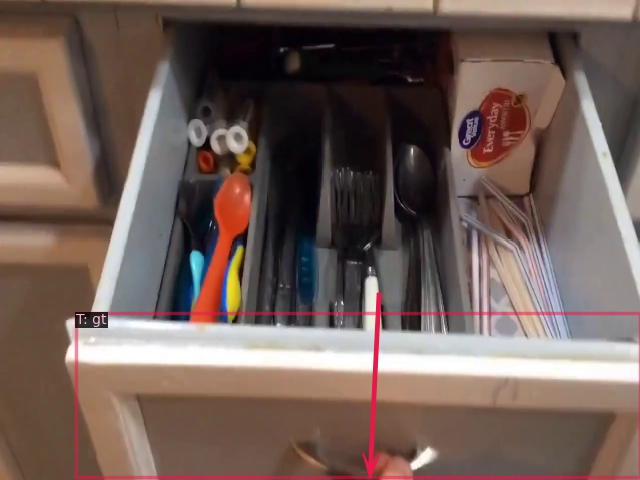}}\\

    \frame{\includegraphics[width=0.15\linewidth]{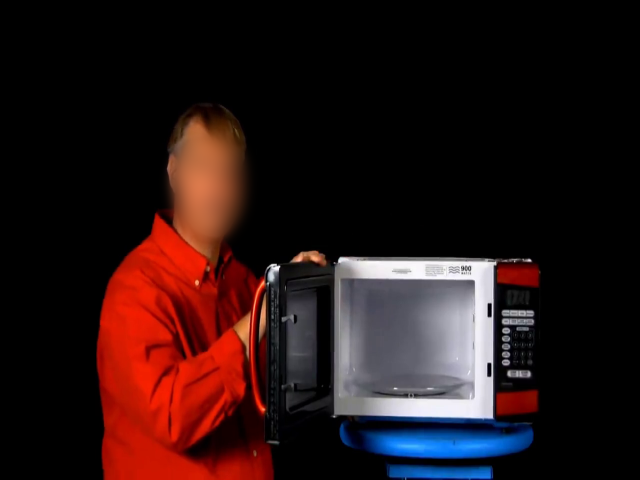}}
    & \frame{\includegraphics[width=0.15\linewidth]{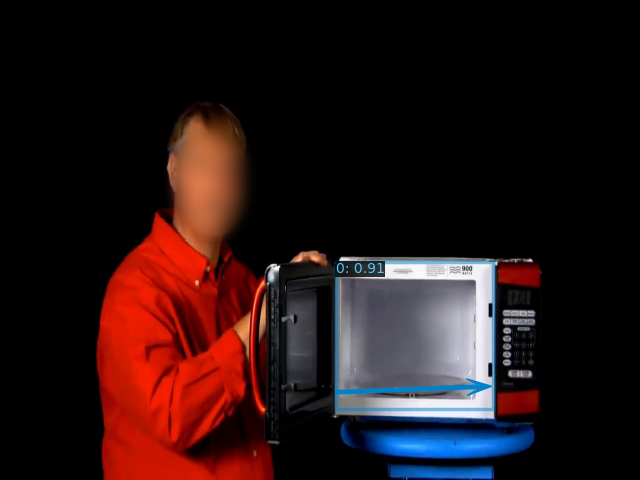}}
    
    & \frame{\includegraphics[width=0.15\linewidth]{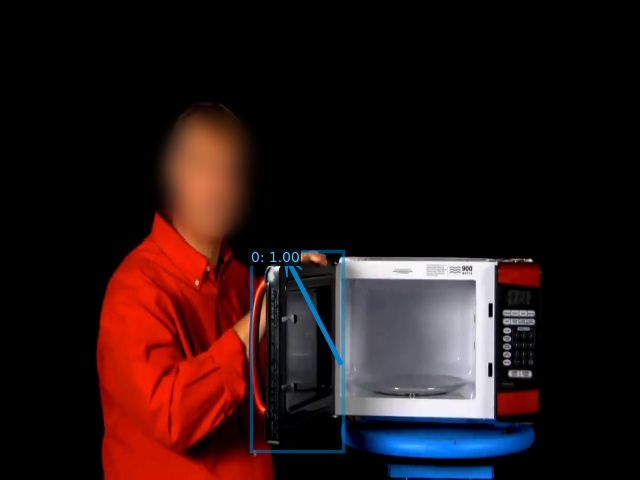}}
    & \frame{\includegraphics[width=0.15\linewidth]{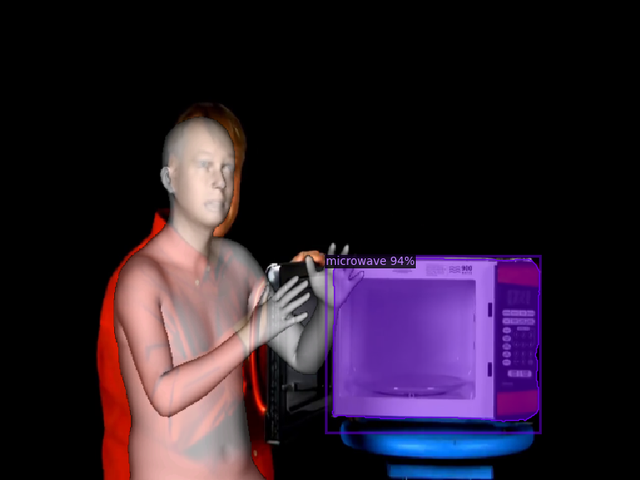}}
    
    & \frame{\includegraphics[width=0.15\linewidth]{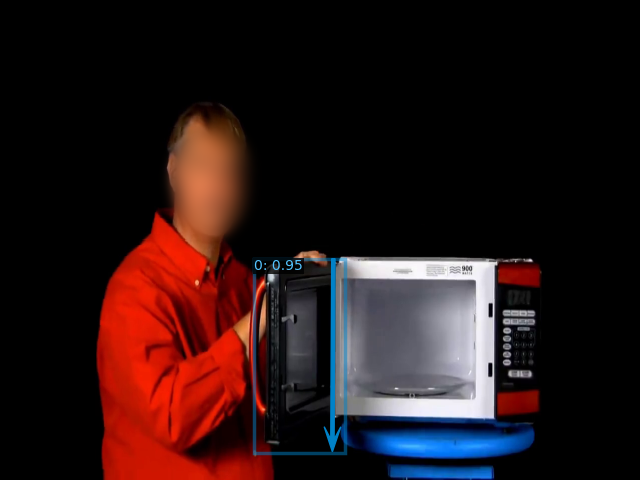}}
    & \frame{\includegraphics[width=0.15\linewidth]{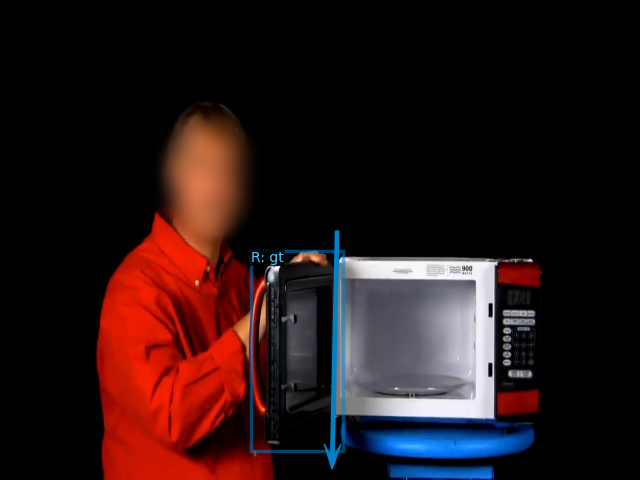}}\\
    
    \bottomrule
    \end{tabular}
    \caption{We compare our approach with four baselines. See detailed discussions in the text. We show translation in \textbf{\textcolor{AccessibleRed}{Pink}} and rotation in \textbf{\textcolor{AccessibleBlue}{Blue}}, except D3D-HOI which uses a different detector. }
    \vspace{-1em}
    \label{fig:detection_wall}
\end{figure*}

We have introduced a method that can infer 3D articulation in Section \ref{sec:approach}.
In the experiments, we aim to answer the following questions: (1) how well can one detect 3D articulating objects from ordinary videos; (2) how well do alternate approaches to the problem do? 

\subsection{Experimental Setup} 

We first describe the setup of our experiments. Our method aims to look at an ordinary RGB video and infer information about an articulated plane on a object in 3D, including: whether the object is articulating, its extent, and the projection of its rotation or translation axis. We therefore evaluate our approach on two challenging datasets, using metrics that capture various aspects of a 3D plane articulating in 3D.

\bfpar{Datasets:} We validate our approach on both Internet videos (described in Section~\ref{sec:dataset}) and the Charades dataset \cite{sigurdsson2016hollywood}. We use Charades for cross-dataset evaluations. We focus on Charades videos that are opening objects (doors, refrigerators, etc.), and annotate 2491 frames across 479 videos; we also randomly sample 479 negative videos containing 4401 negative frames. Our Charades annotation process is similar to Internet videos, with the exceptions that: we annotate only rotations as Charades contains few translation articulations; and we do not annotate surface normals.

\bfpar{Evaluation Criteria:} 
Evaluation of our approach is non-trivial, since our assumed input (RGB videos) precludes measuring outputs quantitatively in 3D. We therefore evaluate our approach on a series of subsets of the problem. We stress from the start though that these metrics are what can be measured (due to the use of RGB inputs), as opposed to the full rich output. 

\noindent {\it Articulation Recognition:} 
We first independently evaluate the ability to detect whether  someone is articulating this object at a point in time. We frame this as a binary prediction problem. This is surprisingly difficult in real scenes because objects are typically partially occluded by humans when humans articulate them, and because humans often touch a articulated objects (e.g., cleaning the surface) without opening it.
We use AUROC to measure the performance.

\noindent {\it Articulation Description:} 
We next evaluate the ability of a system to detect the articulated object, corresponding articulation type (rotation/translation), axes, and surface normals.
We follow other approaches \cite{Qian2020,tulsiani2018factoring,kulkarni20193d,nie2020total3dunderstanding,jin2021planar} that reconstruct the scene factored into components and treat it as a 3D detection problem, evaluated using average precision (AP).
We define error metrics as follows:
({\it Bounding box}) IoU, thresholded as 0.5. We find the normal COCO AP, which measures IoU up to 0.95, to be too strict because the precise boundaries of articulating parts are often occluded by people and hard to annotate. ({\it Axes}) EA-score from the semantic line detection literature~\cite{zhao2021deep}. This metric handles a number of edge cases; we use $0.5$ as the threshold as recommended by \cite{zhao2021deep}. 
({\it Surface normal}) mean angle error, thresholded at 30$^\circ$, following \cite{wang2015designing,eigen2015predicting}.
A prediction is a true positive only if all errors are lower than our thresholds. We calculate the precision-recall curve based on that and report AP for varying combinations of metrics.

\bfpar{Baselines:} 
Prior approaches for articulation detection have focused on robots, synthetic datasets, and real-world RGBD scans. These are different from our setting for two reasons. First, videos of people articulating objects show a noisy background with a person interacting with and occluding the object, as opposed to an isolated articulated object in a simulator. Second, RGB videos do not have depth, which is often a requirement of existing articulation models. For example~\cite{li2020category} requires depth, and while they show results on real-world depth scans, their RGBD scans only contain a static object without humans.
We propose to compare with the following methods.

\begin{table*}
   \centering
   \caption{We report AUROC for Articulation Recognition, as well as AP for Articulation Description. To separate out difficulties in detecting articulation and characterizing its parameters, we assist Flow+Normal and 3DADN+SAPIEN with ground truth bounding box and denote it as gtbox. 3DADN+SAPIEN cannot detect most objects without the help of gtbox.}
   \label{tab:main}
   \scalebox{0.87}{
   \begin{tabular}{lcccccccc}
      \toprule
      & & Recog. & \multicolumn{3}{c}{Rotation} & \multicolumn{3}{c}{Translation}\\
      Methods & gtbox & AUROC & bbox & bbox+axis & bbox+axis+normal & bbox & bbox+axis & bbox+axis+normal\\
      \midrule
      Flow \cite{teed2020raft} + Normal \cite{chen2020oasis} & \msredcross & 68.5 & 7.7 & 0.3 & 0.0 & 0.3 & 0.0 & 0.0 \\
      Flow \cite{teed2020raft} + Normal \cite{chen2020oasis} & \dgreencheck & - & - & 3.0 & 0.3 & - & 1.4 & 0.7 \\
      D3D-HOI~\cite{xu2021d3dhoi} Upper Bound & \msredcross & 62.7 & 28.8 & 19.7 & n/a & 4.70 & 4.7 & n/a \\
      3DADN + SAPIEN~\cite{xiang2020sapien} & \dgreencheck & - & - & 16.8 & 1.40 & - & 15.1 & 0.40 \\
      \textbf{Ours} & \msredcross & \bf{76.6} & \bf{61.3} & \bf{30.4} & \bf{17.2} & \bf{34.0} & \bf{27.1} & \bf{17.9}\\
      \bottomrule
   \end{tabular}
   } %
   \vspace{-1em}
\end{table*}

\vspace{1mm} \noindent 
{\it 3DADN + SAPIEN~\cite{xiang2020sapien} Data:} 
To test whether we can solve the problem just by training on synthetic data, we create a synthetic data-based method where we train our 3DADN system on synthetic data. %
We render a synthetic dataset using SAPIEN~\cite{xiang2020sapien} by randomly sampling and driving 3D objects. 
We filtered 1053 objects of 18 categories with movable planes from PartNet-Mobility Dataset~\cite{xiang2020sapien}, such as doors and laptops. We render frames with the objects articulated, with location parameters picked to give plausible scenes, and extract the information needed to train 3DADN. 
Without a background, the detection problem becomes trivial, so to mimic real 3D scenes, we blend the renderings with random ScanNet~\cite{dai2017scannet} images as the background and render synthetic humans from SURREAL~\cite{varol17_surreal}. For fair comparison, we use the same ScanNet+SURREAL images used to train our system's plane parameter head. When evaluated on SAPIEN data, this approach performs well and obtains an AP of (bbox) 60.3, (bbox+rot) 64.1, (bbox+rot+normal) 41.0.

\vspace{1mm}
\noindent 
{\it Bottom-up Optical Flow~\cite{teed2020raft} and Surface Normal Change~\cite{chen2020oasis} (Flow+Normal):} To test whether the data can be solved by the use of fairly simple cues, we construct a baseline that uses Optical Flow~\cite{teed2020raft} (since articulating objects tend to cause movement) and Surface Normals~\cite{chen2020oasis} (since rotating planes change their orientation). Both flow and normals provide a $H \times W$ map that can be analyzed. We also use the output of a human segmentation system~\cite{iglovikov2018ternausnet} that was trained on multiple datasets and mask normal and flow magnitude maps wherever it improves performance. Given these maps, we recognize the presence of articulation via logistic regression on a feature vector consisting of the fraction of pixels above multiple thresholds; we recognize bounding boxes via thresholding and finding the tightest enclosing box; we estimate rotation axis as perpendicular to the mean flow change in the bounding box (flow tends to increase away from hinges); we find translation axis using mean flow direction in the box; we find articulation normal using mean predicted normals in the box. Throughout, we use the optimal option of surface normals and flow; this hybrid system performs substantially better than either flow or normals alone.

\vspace{1mm}
\noindent {\it Baselines with + GT Box:} 
To separate out difficulties in detecting articulation and characterizing its parameters, we also experiment with giving baselines ground-truth bounding box information about the articulating object. This gives an upper-bound on performance.

\vspace{1mm}
\noindent
{\it D3D-HOI \cite{xu2021d3dhoi} Upper Bound: } 
We compare with D3D-HOI since it accepts RGB video as input and detects how humans articulate objects.
A direct comparison with D3D-HOI is challenging since it only works when EFT \cite{joo2020exemplar} reconstructs 3D human poses and Pointrend \cite{kirillov2020pointrend} detects the objects that are assumed to articulate and correct CAD models are chosen for the object.
However, EFT does not work well on the dataset due to truncated or multiple humans on Internet videos \cite{rockwell2020full,Kocabas2021}.
We therefore report upper-bounds on the performance.
We assume it predicts the ground truth bounding box, when EFT mask and pseudo ground truth 2D human segmentation mask~\cite{TernausHuman} has IoU $>$ 0.5 and PointRend \cite{kirillov2020pointrend} produces a mask on articulated objects with confidence $>$ 0.7.

\noindent {\it Ours:}
This is our proposed method. It includes both the per-frame approach described in Section~\ref{sec:approach_detect} and the optimization approach of Section~\ref{sec:approach_optimize}. We note that this approach also produces outputs that are not being quantitatively measured, such as a 3D plane articulating in 3D. These are qualitatively shown in Figures~\ref{fig:teaser} and \ref{fig:vis_wall}.

\subsection{Results}

We first show qualitative results in Figure \ref{fig:vis_wall}.
On challenging Internet videos, our approach usually detects and recovers the 3D articulated plane regardless of categories.

In Figure~\ref{fig:detection_wall}, we compare our approach with four baselines visually.
Flow can occasionally locate articulation (third row), but in most cases, flow is not localized to only the object articulating (e.g. camera movement, top row). Training purely on SAPIEN~\cite{xiang2020sapien} data has difficulty detecting articulated objects in Internet videos, even if we show all detected objects with confidence score $> 0.1$. It learns some information of articulation axes when we assist it with ground truth bounding boxes. D3D-HOI~\cite{xu2021d3dhoi} relies on both EFT~\cite{joo2020exemplar} to detect humans and PointRend~\cite{kirillov2020pointrend} to detect objects. However, EFT has diffculty predicting 3D humans on Internet videos.

\bfpar{Quantitative Results.}
We evaluate the approach quantitatively on the three tasks in Table~\ref{tab:main}. Our approach substantially outperforms the alternate methods. While statistically-combined bottom-up cues~\cite{teed2020raft,chen2020oasis} do better than chance at predicting the presence of an articulation, they are substantially worse than the proposed approach and fail to obtain sensible bounding boxes. Even when given the ground-truth box, this method fails to obtain good axes. Due to the frequency of truncated humans in Internet videos~\cite{rockwell2020full,Kocabas2021}, D3D-HOI~\cite{xu2021d3dhoi}'s performance upper-bound is substantially lower than our method's performance. The detection system when trained on synthetic data from~\cite{xiang2020sapien} fails on our system; when given a good bounding box, synthetic training data obtains reasonable, but inferior numbers and poor accuracy in predicting normals. 

\bfpar{Ablations -- Optimization.} Our optimization produces modest gains in recognition accuracy and axis localization in 2D: It improves recognition AUROC from 74.0 to 76.6, rotation AP from 16.6 to 17.2 and translation AP from 14.3 to 17.9.
This small gain is understandable because the evaluation is per-frame and the optimization mainly seeks to make the predictions more consistent. If we quantify the consistency in the results before and after optimization, we find that the EAScore~\cite{zhao2021deep} between tracked predicted frames increases from 0.69 (before optimization) to 0.96 (after optimization).

\begin{figure}
    \centering
    \begin{tabular}{c@{\hskip4pt}c@{\hskip4pt}c@{\hskip4pt}c@{\hskip4pt}c@{\hskip4pt}c}
    \toprule
    Pred & GT & Pred & GT & Pred & GT \\
    \midrule
    \frame{\includegraphics[width=0.14\linewidth]{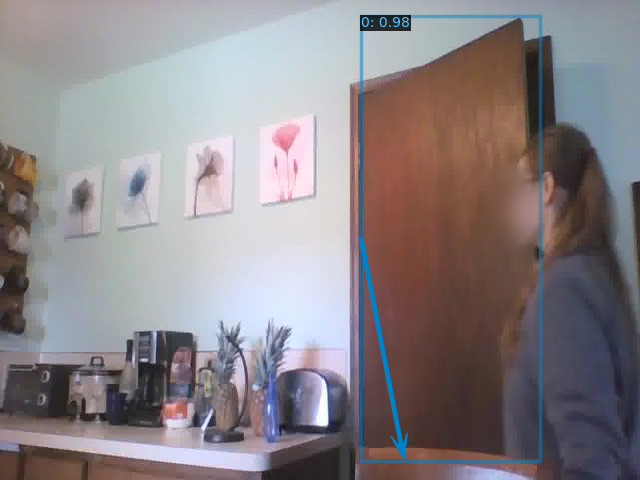}}
    & \frame{\includegraphics[width=0.14\linewidth]{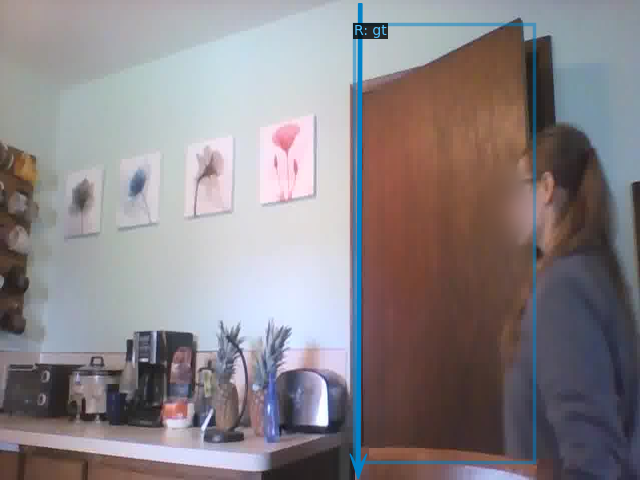}}
    & \frame{\includegraphics[width=0.14\linewidth]{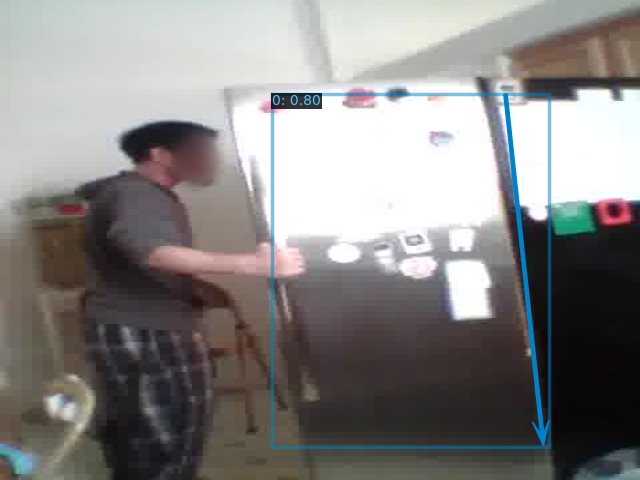}}
    & \frame{\includegraphics[width=0.14\linewidth]{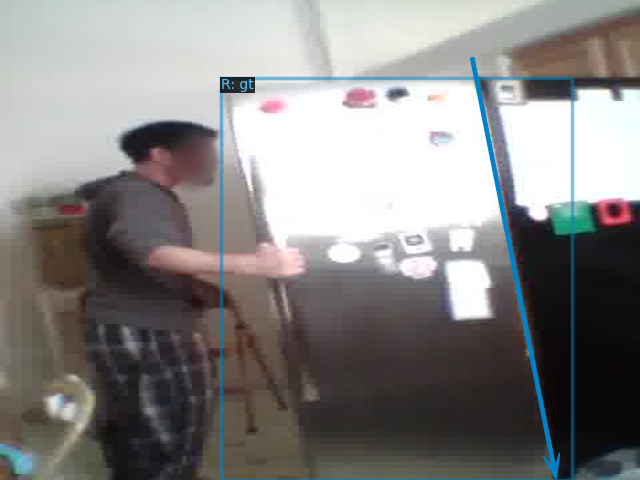}}
    & \frame{\includegraphics[width=0.14\linewidth]{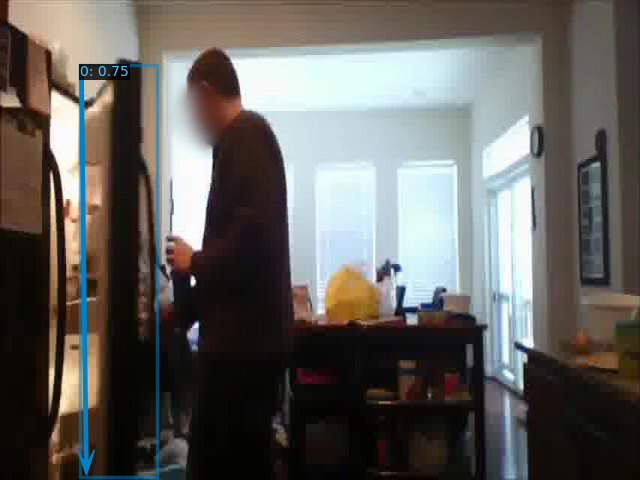}}
    & \frame{\includegraphics[width=0.14\linewidth]{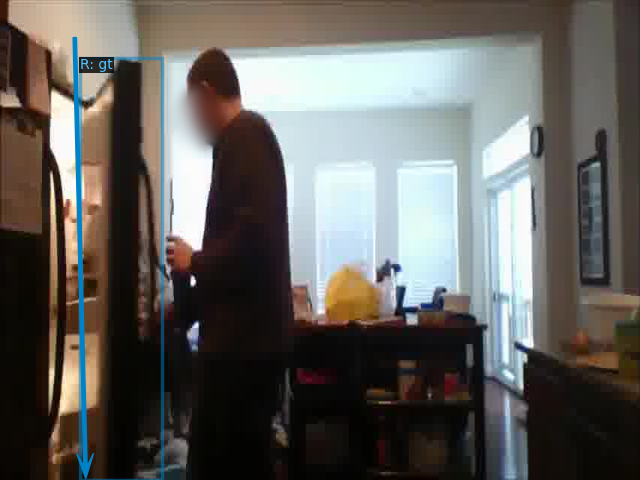}}\\
    
    \frame{\includegraphics[width=0.14\linewidth]{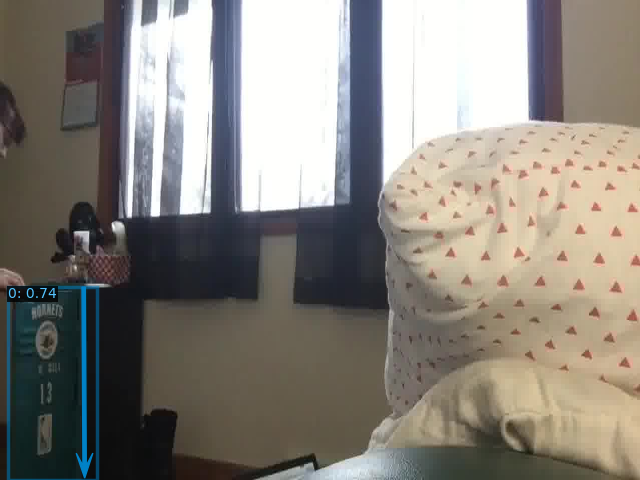}}
    & \frame{\includegraphics[width=0.14\linewidth]{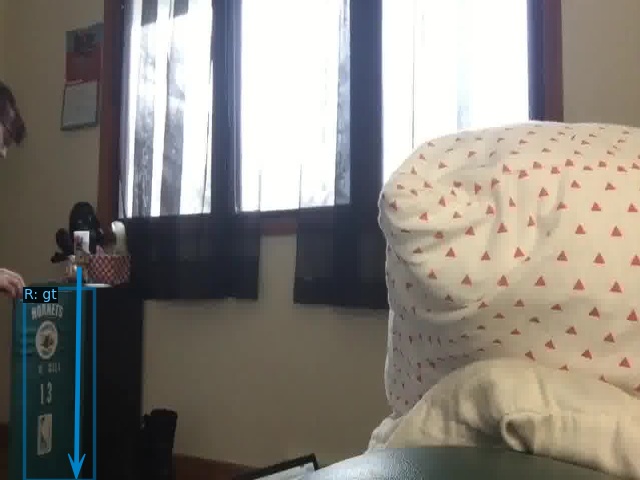}}
    & \frame{\includegraphics[width=0.14\linewidth]{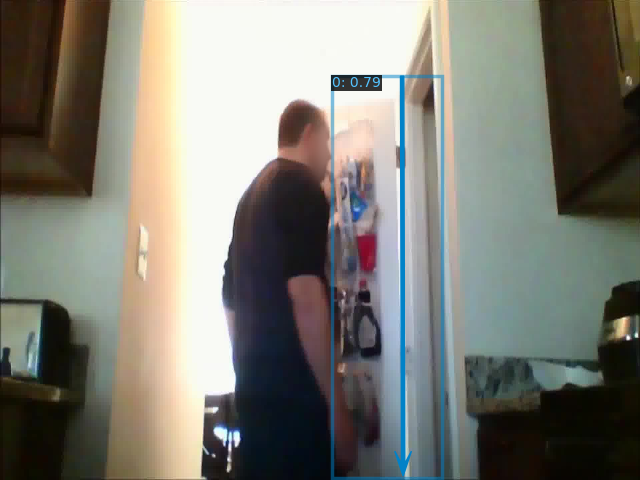}}
    & \frame{\includegraphics[width=0.14\linewidth]{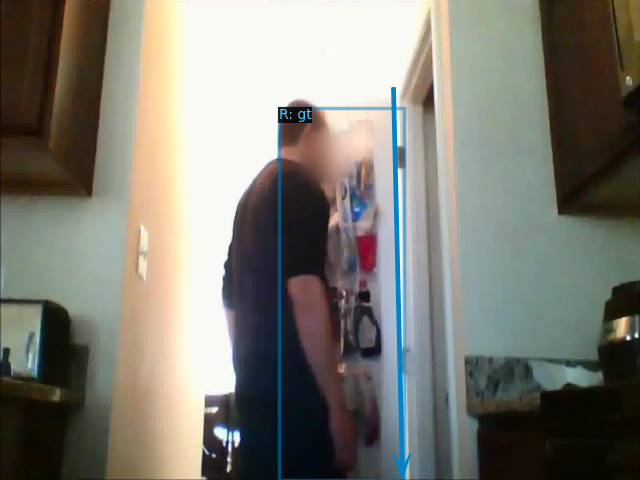}}
    & \frame{\includegraphics[width=0.14\linewidth]{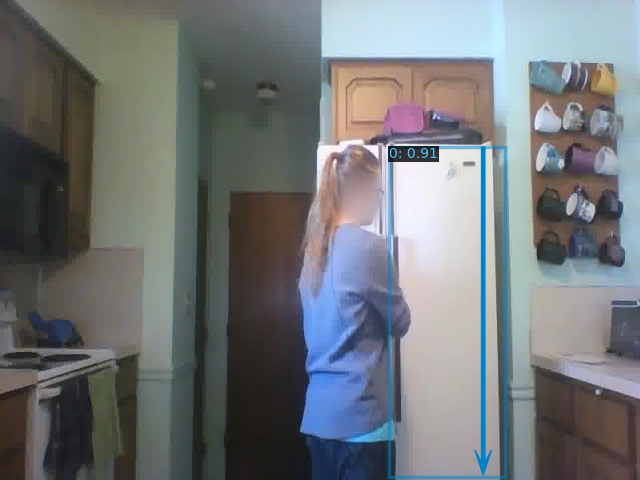}}
    & \frame{\includegraphics[width=0.14\linewidth]{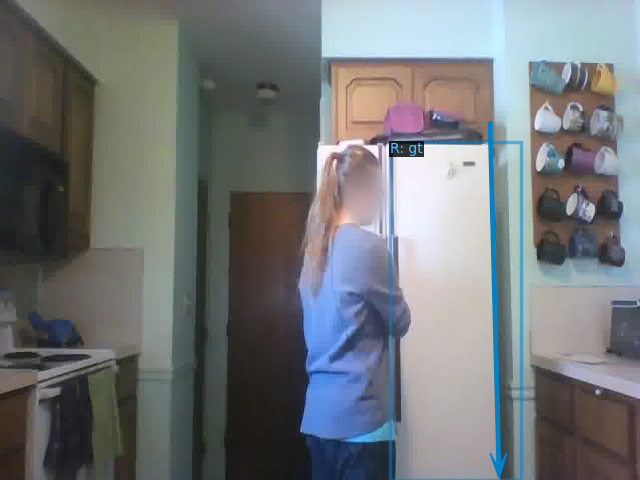}}\\
    \bottomrule
    \end{tabular}
    \caption{Qualitative results on Charades dataset. Without finetuning on Charades data, our model obtains strong performance on detecting and characterizing 3D articulation.}
    \label{fig:results_charades}
\end{figure}

\begin{table}
   \centering
   \caption{Evaluation on Charades dataset \cite{sigurdsson2016hollywood}. We only report rotation AP since Charades does not have sufficient translation motion.}
   \label{tab:charades}
   \scalebox{0.81}{
   \begin{tabular}{l@{\hskip20pt}cc@{\hskip20pt}ccccc}
      \toprule
      & & Recog. & \multicolumn{2}{c}{Rotation}\\
      Methods & gtbox & AUROC & bbox & bbox+axis \\
      \midrule
      Flow \cite{teed2020raft} + Normal \cite{chen2020oasis} & \msredcross & 53.7 & 3.1 & 0.2 \\
      Flow \cite{teed2020raft} + Normal \cite{chen2020oasis} & \dgreencheck & - & - & 4.2 \\
      D3D-HOI Upper Bound & \msredcross & 55.9 & \bf{14.9} & \bf{13.7}\\
      3DADN + SAPIEN~\cite{xiang2020sapien} & \dgreencheck & - & - & 1.54\\
      \textbf{Ours} & \msredcross &  \textbf{58.4} &  12.0 & 12.8 \\
      \bottomrule
   \end{tabular}
   } %
   \vspace{-1.5em}
\end{table}

\subsection{Generalization Results}

We next test our trained models {\it without fine-tuning} on Charades~\cite{Sigurdsson2016}. We show
results in Figure \ref{fig:results_charades}.
Our approach typically generates reasonable estimations. We find that
the video quality and resolution of Charades is lower relative to our videos, with 
many dark or blurry videos. 

We also show quantitative evaluations in Table \ref{tab:charades}. Here, our performance is slightly diminished. However, we substantially outperform the baselines. We are only marginally outperformed by D3D-HOI upper bound, which assumes perfect performance so long as the data can be obtained.

\begin{figure}
    \centering
    \begin{tabular}{c@{\hskip4pt}c@{\hskip4pt}c@{\hskip4pt}c@{\hskip4pt}c@{\hskip4pt}c}
    \toprule
    Pred & GT & Pred & GT & Pred & GT   \\
    \midrule
    \frame{\includegraphics[width=0.14\linewidth]{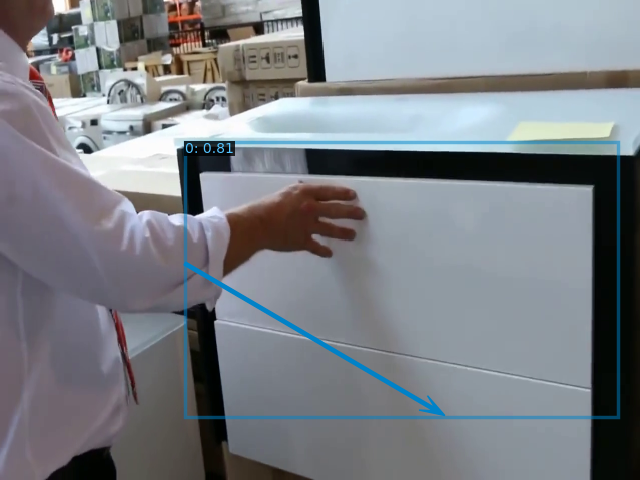}}
    & \frame{\includegraphics[width=0.14\linewidth]{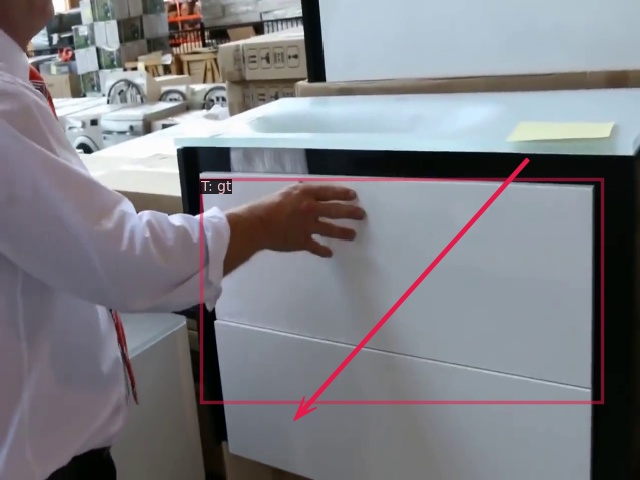}}
    & \frame{\includegraphics[width=0.14\linewidth]{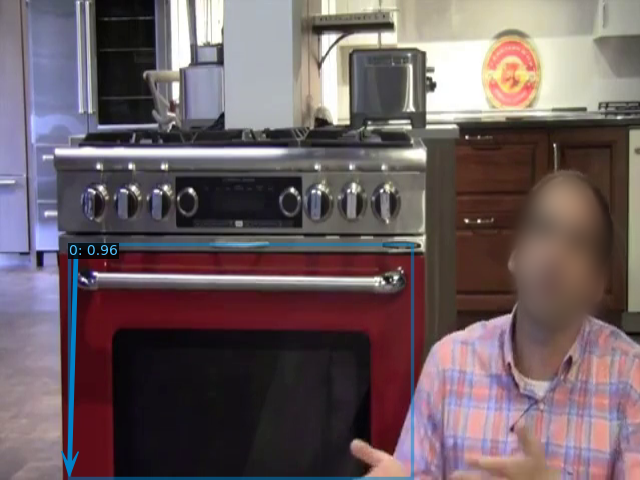}}
    & \frame{\includegraphics[width=0.14\linewidth]{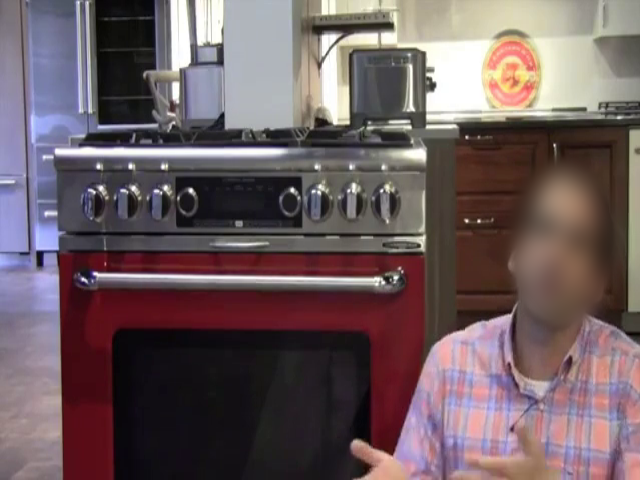}}
    & \frame{\includegraphics[width=0.14\linewidth]{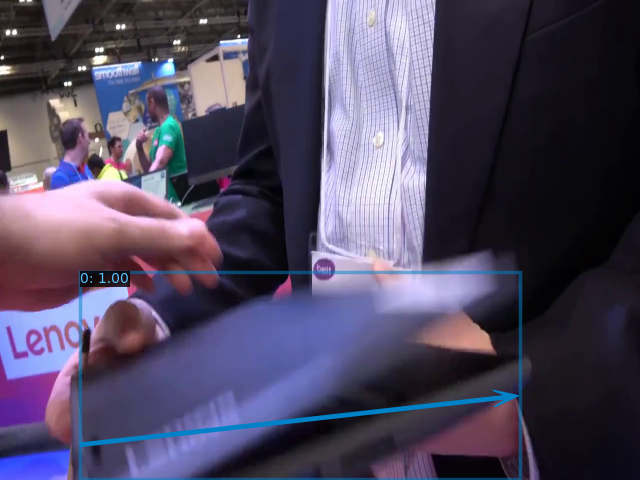}}
    & \frame{\includegraphics[width=0.14\linewidth]{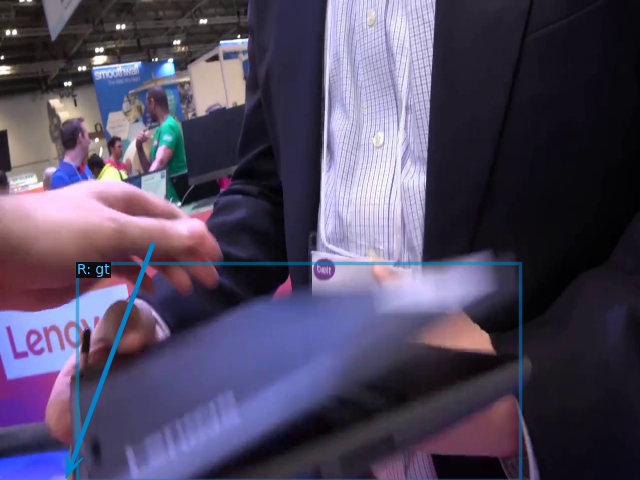}}\\
    \frame{\includegraphics[width=0.14\linewidth]{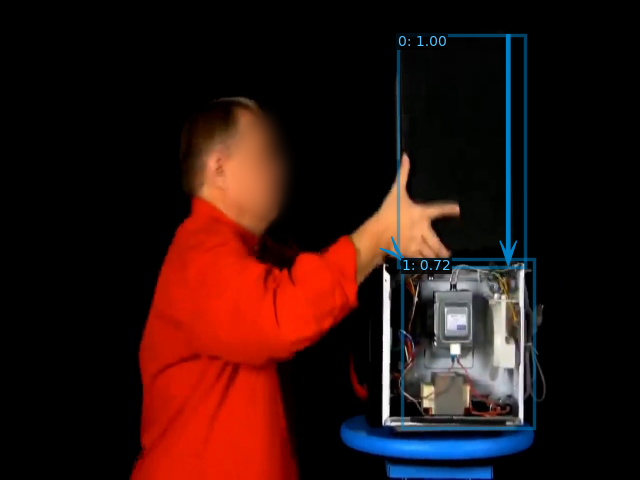}}
    & \frame{\includegraphics[width=0.14\linewidth]{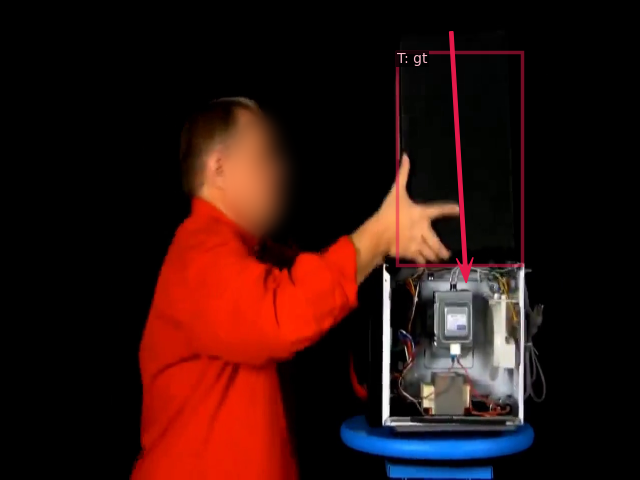}}
    &\frame{\includegraphics[width=0.14\linewidth]{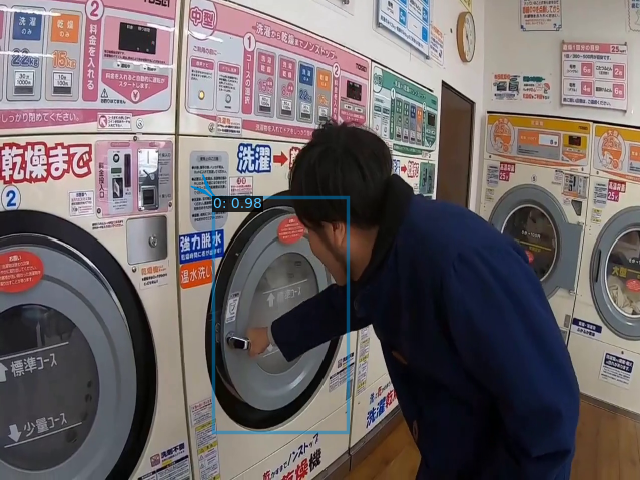}}
    & \frame{\includegraphics[width=0.14\linewidth]{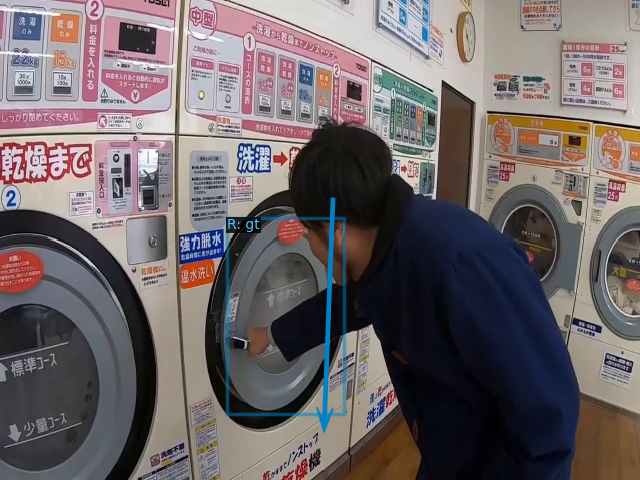}}
    & \frame{\includegraphics[width=0.14\linewidth]{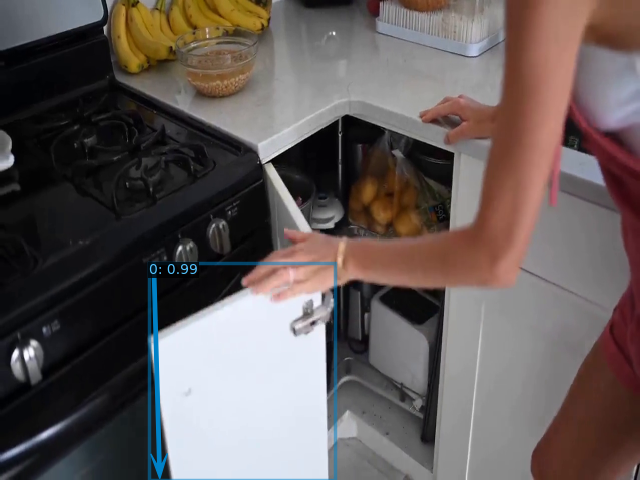}}
    & \frame{\includegraphics[width=0.14\linewidth]{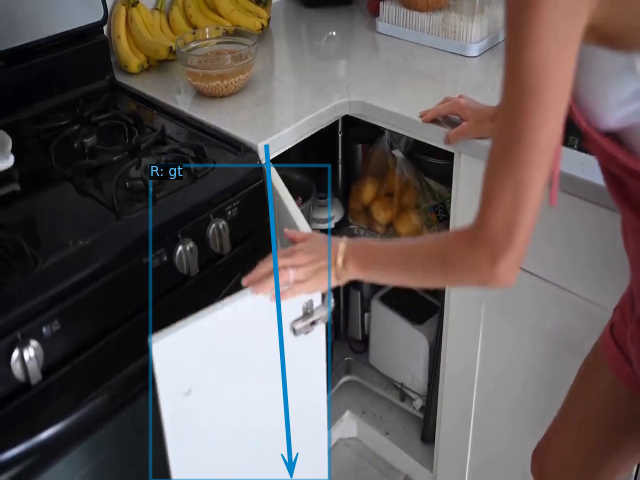}}\\
    \bottomrule
    \end{tabular}
    \caption{Typical failure modes. (1) Ambiguity of articulation type; (2) Axis outside of the frame or ambiguity of articulation axis location due to symmetry; (3) Object has complex motions (a person moving an object while articulating it; the rotation axis is outside of the articulating surface). }
    \label{fig:failure}
    \vspace{-1.5em}
\end{figure}

\subsection{Limitations and Failure Modes}
\label{sec:failure}

We finally discuss our limitations and typical failure modes in Figure \ref{fig:failure}.
We find some examples are particularly challenging:
(1) Column 1: some images may contain hard examples where the axis types are hard to figure out.  (2) Column 2: the axis is outside of the image frame or its location is ambiguous due to symmetry or occlusion. (3) Column 3: the object has complex dynamics or dual axes; for example, a person moving a laptop while opening it or the cabinet has multiple joints.

\section{Conclusion}
\label{sec:conclusion}

We have demonstrated our approach's ability to detect and characterize 3D planar articulation of objects from ordinary videos.
Future work includes combining 3D shape reconstruction with the articulation detection pipeline. 

Our approach can have positive impacts by helping build smart robots that are able to understand and manipulate articulated objects.
On the other hand, our approach may be useful for surveillance activities. Moreover, our network is trained on Internet videos and deep networks may amplify biases in the data.

\vspace{0.75em}
\noindent
{\bf Acknowledgments} 
This work was supported by the DARPA Machine Common Sense Program and Toyota Research Institute.
Toyota Research Institute (``TRI'') provided funds
to assist the authors with their research but this article solely
reflects the opinions and conclusions of its authors and not
TRI or any other Toyota entity.
We thank Dandan Shan, Jiaqi Geng, Sarah Jabbour and Ruiyu Li for their help with data collection, Mohamed El Banani for his help of blender, Fanbo Xiang for his help of SAPIEN, as well as Yichen Yang and Ziyang Chen for their help of Figure 2.
We also thank Justin Johnson, Jiteng Mu, Tiange Luo and Max Smith for helpful discussions.

{\small
\bibliographystyle{ieee_fullname}
\bibliography{local}
}

\clearpage
\appendix

\section{Implementation Details}

\begin{table*}[h!]
    \centering
    \caption{Overall architecture for our proposed network. The backbone, RPN and plane branches are identical to \cite{jin2021planar}. 
    The RPN predicts a bounding box for each of $A$ anchors in the input feature map. $C$ is the number of categories (here = 2 for rotation and translation). We use class agnostic mask because the mask head is trained on ScanNet.
    TConv is a transpose convolution with stride 2. ReLU is used between all Linear, Conv and TConv operations.
    Depth branch uses Conv and Deconv layers to generate a depthmap with the same resolution as the input image.
    }
    \resizebox{\textwidth}{!}{
        \begin{tabular}{cclc}
            \toprule
            Index & Inputs & Operation & Output shape \\
            \midrule
            (1)	&	Inputs	&	Input Image	&	$H\times W\times3$	\\
            (2)	&	(1)	&	Backbone: ResNet50-FPN	&	$h\times w\times 256$	\\
            (3)	&	(2)	&	RPN	&	$h\times w\times A\times4$	\\
            (4)	&	(2),(3)	&	\texttt{RoIAlign}	&	$14\times14\times256$	\\
            (5)	&	(4)	&	Box: 2$\times$downsample, Flatten, $\mathrm{Linear}(7\times 7\times 256\rightarrow 1024)$, $\mathrm{Linear}(1024\rightarrow5C)$	&	$C\times5$	\\
            (6)	&	(4)	&	Mask: $4\times\mathrm{Conv}(256\rightarrow 256,3\times 3)$, $\mathrm{TConv}(256\rightarrow 256,2\times 2,2)$, $\mathrm{Conv}(256\rightarrow 1\times 1)$	&	$1\times28\times 28$	\\
            (7)	&	(4)	&	Normal: $4\times\mathrm{Conv}(256\rightarrow 256,3\times3)$, $\mathrm{Linear}(14\times14\times256\rightarrow1024)$, $\mathrm{Linear}(1024\rightarrow3)$	&	$1\times3$	\\
            (8)	&	(4)	&	Rotation Axis: $4\times\mathrm{Conv}(256\rightarrow 256,3\times3)$, $\mathrm{Linear}(14\times14\times256\rightarrow1024)$, $\mathrm{Linear}(1024\rightarrow3)$	&	$C\times3$	\\
            (9)	&	(4)	&	Translation Axis: $4\times\mathrm{Conv}(256\rightarrow 256,3\times3)$, $\mathrm{Linear}(14\times14\times256\rightarrow1024)$, $\mathrm{Linear}(1024\rightarrow2)$	&	$C\times2$	\\
            (10)	&	(2)	&	Depth	&	$H\times W\times 1$	\\
            \bottomrule
        \end{tabular}
    }
    \label{tab:full_arch}
\end{table*}

\noindent
\textbf{Detector.}
Our network architecture is shown in Table~\ref{tab:full_arch}. We use Detectron2 \cite{wu2019detectron2} to implement our articulation detection and take the codebase from SparsePlanes \cite{jin2021planar}. Our articulation head predicts rotation and translation axis separately. Each of the branch takes the RoI feature from the backbone and uses four convolutional layers with 256 channels and two linear layers to regress the axes. The rotation branch predicts a three dimensional vector and the translation branch predicts a two dimensional vector.
While they train the model on Matterport3D \cite{chang2017matterport3d}, we start from COCO pretraining and train the model on our Internet videos.
The training is run on a single RTX 2080Ti.
 
\vspace{0.75em}
\noindent
\textbf{Temporal Optimization.}
For tracking, we use $0.5$ as our IoU threshold. For articulation model fitting, we use the ScanNet camera intrinsics as our assumed camera intrisics, since the plane and depth heads of the model is trained on ScanNet \cite{dai2017scannet}.
We use PyTorch to implement the temporal optimization.
For 3D transformations we use PyTorch3D \cite{ravi2020pytorch3d} so that it is compatible.
The optimization is parallel and runs on 8 GTX 1080Ti gpus of an internal cluster to save inference time and it can be run on a single gpu.

\section{Data Collection Pipeline}

Our semi-automatic data collection consists of an automatic pipeline to download and filter Internet videos to remove clear negatives, and a manual annotation to label articulated objects.
We also discuss how we filter Charades dataset \cite{sigurdsson2016hollywood} since it involves slightly different steps.

\subsection{Filtering Internet videos}

\noindent
{\bf Youtube queries.}
We start with $10$ initial common articulated objects in our daily life: door, laptop, oven, refrigerator, washing machine, dishwasher, microwave oven, drawer, cabinet and box.
Using the combination of words, we make a list of queries for each initial category, e.g. ``best laundry tips'', following 100DOH \cite{Shan20}.
To improve the number of videos we can find on the Internet and the diversity of the dataset, we also translate the queries into Chinese, Japanese, Hindi and Spanish.
We search these queries on Youtube and download related Creative Commons videos.

\vspace{0.75em}
\noindent
{\bf Converting videos to shots.}
Within these videos, we find stationary continuous shots by fitting homographies \cite{Hartley04} on ORB \cite{rublee2011orb} features.
In practice, we find ORB \cite{rublee2011orb} are much faster and slightly more robust than SIFT \cite{lowe2004distinctive} features, so it saves a lot of computing time.

\vspace{0.75em}
\noindent
{\bf Filtering videos based on interaction.}
A lot of stationary shots does not contain any people or objects of interest.
We further filter out video shots based on a hand interaction detector trained on 100$K$+ frames of Internet data \cite{Shan20}.
In practice, we find it works well and we believe it is because \cite{Shan20} is also trained on Internet data.
For each video shot, we run the hand interaction detector on frames evenly sampled every 1 second. 
We remove video shots which do not have hand interactions at all.

\vspace{0.75em}
\noindent
{\bf Filtering categories of interests.}
We further filter object of interests by an object detector trained on COCO \cite{Lin2014} and LVIS \cite{gupta2019lvis}. 
We use the pretrained model of Faster R-CNN (X101-FPN, 3x) from Detectron2 \cite{wu2019detectron2}.
For categories that COCO does not have annotation (e.g. washing machines), we use LVIS since it has much more categories.
Especially, LVIS does not have annotations for doors, and we use doorknob instead.

\subsection{Annotating Articulated Objects}

Finally, we use annotate articulated objects using crowdsourcing. 
We split video shots into 3s clips, where the fps is 30.
Therefore, there are 90 frames per clip.

We use Hive \footnote{\url{https://thehive.ai/}} as our data annotation platform.
We include the screenshot and estimated hourly pay for each step, since these steps are separate from each other.
The hourly pay is a rough estimation since we only have limited worker statistics provided by Hive and we do not manage it ourselves.

\begin{figure*}[h]
    \centering
    \includegraphics[width=0.9\linewidth]{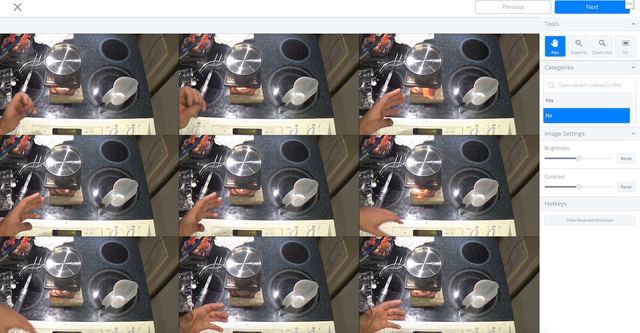}
    \caption{The screenshot of recognize articulated clips.}
    \label{fig:supp_anno_binary}
\end{figure*}

\vspace{0.75em}
\noindent
\textbf{Recognizing articulated clips.}
The first step is to judge if the video clips have objects which are being articulated. 
This is a binary classification question.
We show 9 key frames of the video shot (sample every 10 frames) and ask workers to classify.
The screenshot is shown in Figure \ref{fig:supp_anno_binary}. 
We pay \$$0.015$ for each clip, each clip is annotated by about $2$ workers (which is managed by Hive based on consensus and out of our control), and we estimate they can annotate $8$ clips per minute.
Therefore, the estimated hourly pay is $0.015 \cdot 8 \cdot 60 / 2 = \$3.6$.

\begin{figure*}[h]
    \centering
    \includegraphics[width=0.9\linewidth]{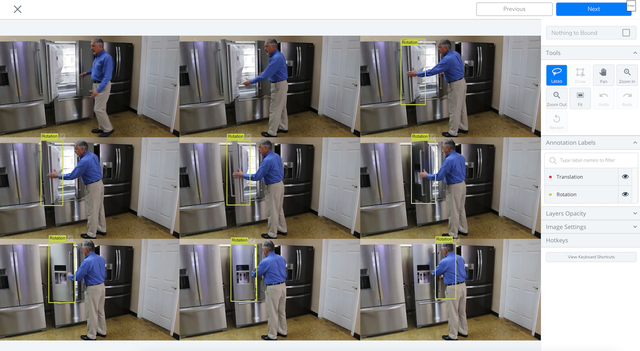}
    \caption{The screenshot of annotating bounding boxes.}
    \label{fig:supp_anno_box}
\end{figure*}

\vspace{0.75em}
\noindent
\textbf{Annotating bounding boxes of articulated objects.}
After labelling positive video clips, we annotate bounding boxes of objects which are being articulated.
We also ask workers to specify the object is being rotated or translated.
We only annotate on 9 key frames of the video clip, since consecutive frames tend to be similar.
The screenshot is shown in Figure \ref{fig:supp_anno_box}.
We pay \$$0.14$ for each clip, each clip is annotated by about $2$ workers, and we estimate they can annotate $1.5$ clips per minute.
Therefore, the estimated hourly pay is $0.14 \cdot 1.5 \cdot 60 / 2 = \$6.3$.

\begin{figure*}[h]
    \centering
    \includegraphics[width=0.9\linewidth]{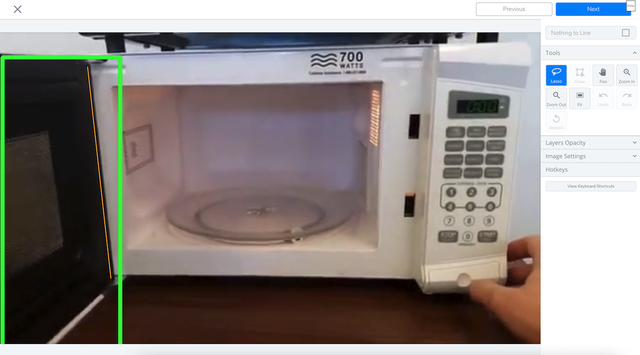}
    \caption{The screenshot of annotating rotation axes.}
    \label{fig:supp_anno_rot}
\end{figure*}

\vspace{0.75em}
\noindent
\textbf{Annotating rotation axis.}
For objects which are annotated to be rotated, we annotate their rotation axes.
We ask workers to draw a line to represent the 2D rotation axis.
The screenshot is shown in Figure \ref{fig:supp_anno_rot}.
We pay \$$0.04$ for each clip, each clip is annotated by about $2$ workers, and we estimate they can annotate $4$ axes per minute.
Therefore, the estimated hourly pay is $0.04 \cdot 4 \cdot 60 / 2 = \$4.8$.

\begin{figure*}[h]
    \centering
    \includegraphics[width=0.9\linewidth]{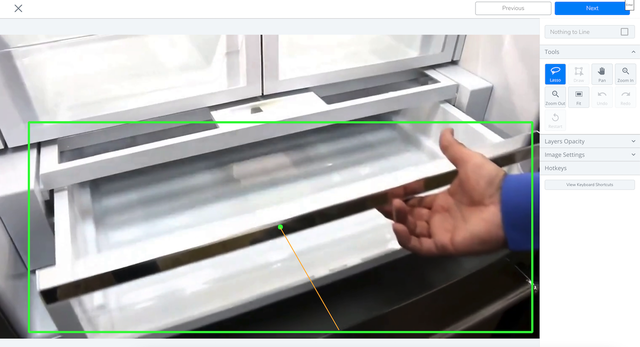}
    \caption{The screenshot of annotating translation axes.}
    \label{fig:supp_anno_trans}
\end{figure*}

\vspace{0.75em}
\noindent
\textbf{Annotating translation axis.}
For objects which are annotated to be translated, we annotate their translation direction.
We also ask workers to draw a line to represent the 2D rotation direction.
However, since translation is only related to the angle of the line and does not need the line offset,
we draw a circle at the center of the bounding box and ask workers to start there.
The screenshot is shown in Figure \ref{fig:supp_anno_trans}.
The estimated hourly pay is \$$4.8$, which is the same as annotating rotation axis since it is defined as the same task ``line segment'' on Hive.

\begin{figure*}[t]
    \centering
    \includegraphics[width=0.9\linewidth]{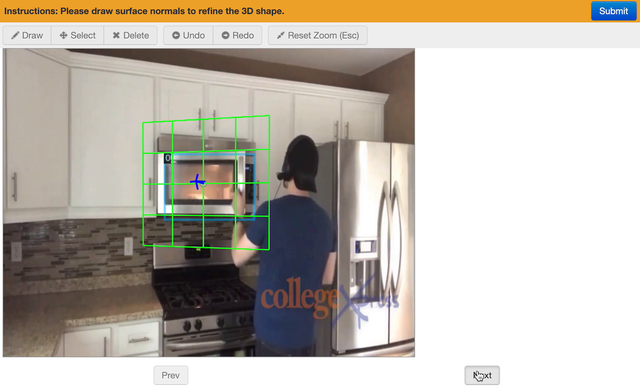}
    \caption{The screenshot of annotating surface normals.}
    \label{fig:supp_anno_normal}
\end{figure*}

\vspace{0.75em}
\noindent
\textbf{Annotating surface normals.}
All bounding boxes and articulation axes are annotated in 2D.
However, in this paper, we are interested in 3D object articulation.
Thus, for the test set, we also annotate the o annotate the surface normal of the plane following~\cite{chen2020oasis}, so we can evaluate how well our model can learn 3D properties.
Since the annotation of surface normals are not available on Hive and we only need surface normals on the test set for evaluation purpose, we do all surface normals annotations ourselves.
The screenshot is show in Figure~\ref{fig:supp_anno_normal}.

Finally, we postprocess the dataset to make sure the dataset does not have offensive content, cartoons, and any videos depicting children.
Since the distribution is unbalanced and negative examples are much more than positive ones,
we only sample a negative clip with hand interaction from the same video shot of positive clips.

\begin{figure*}[t]
    \centering
    \includegraphics[width=\linewidth]{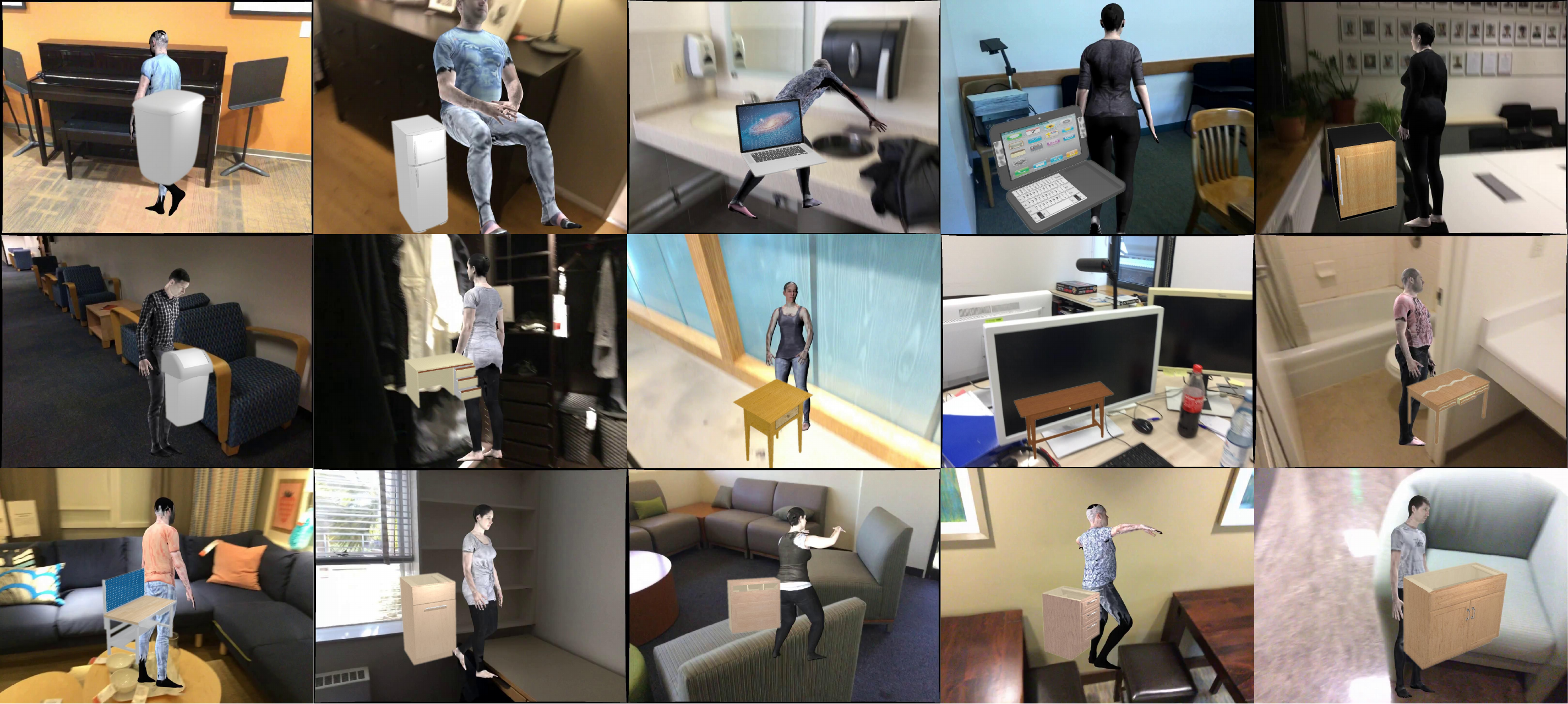}
    \caption{Random examples from Sapien renderings.}
    \label{fig:supp_sapien}
\end{figure*}

\subsection{Annotating Charades Dataset}

We test generalization of our method on Charades without additional training. Data is prepared and annotated using the following process on the original Charades test set.

The Charades test set contains 1863 videos of people acting out designated scripts. Videos are typically short at around 30 seconds. The dataset contains script information and additional annotation, which can be used to filter videos that are highly likely to contain articulation. After initial filtering, we split filtered videos into clips of a pre-defined length. These clips are then annotated as to whether they contain articulation, for articulation bounding box location (if applicable), and articulation angle (if applicable). Annotation is performed using a similar setup to YouTube via Hive. Because Charades is small, we use all clips which Hive workers have labeled as containing articulation.

\noindent
\textbf{Filtering criteria.} Charades contains videos with acting information, which we use to perform filtering. Each video has corresponding categorical actions that can be used to find dense articulation instances. These categorical actions fall into 157 categories such as "Holding some clothes" or "Opening a bag", and are annotated on a one-tenth of a second basis. For example, one given video may have two corresponding actions "Holding some clothes" and "Opening a bag", which correspond to seconds 1.2 - 11.7 and 12.3 - 18.3, correspondingly.
We selecct video clips orresponding to eight categorical actions for our dataset: "Opening a door", "Closing a door", "Opening a laptop", "Closing a laptop", "Opening a closet/cabinet", "Closing a closet/cabinet", "Opening a refrigerator", and "Closing a refrigerator". 

\noindent
\textbf{Gathering filtered clips.} To find corresponding video clips, we first calculate the middle of each action -- e.g. for the 1.2 - 11.7 interval this would be 6.45. Next, we select a 7.5 seconds both greater than and less than the middle of the action. This is subject to beginning and ending of video, and we remove overlapping clips. In our example, "Holding some clothes" has a middle of 6.45 seconds, so the clip would begin at 0 seconds and end at 13.95 seconds. The second clip has a middle of 15.3, and cannot overlap with the first, so would start at 13.95 seconds and end at 22.8. This totals 649 filtered clips of $<=$ 15 seconds.

\noindent
\textbf{Splitting video clips.} Given a set of video clips which correspond to the selected categorical actions, we next split clips into 3 second clips to be used for a standardized articulation framework. Any clip less than 3 seconds is truncated. This results in 2232 3-second mini-clips.

\section{Rendering Sapien data}

In our experiments, we test whether the model trained on synthetic data transfer to Internet videos using Sapien renderings~\cite{xiang2020sapien}.
Here we provide additional details about rendering and example images.
Examples of our renderings are shown in Figure~\ref{fig:supp_sapien}.

To generate these results,
we first randomly sample 3D objects with articulation. 
We filtered 1053 objects of 18 caterories with movable planes from PartNet-Mobility Dataset~\cite{xiang2020sapien}. 18 categories are: ``Box'', ``Dishwasher'', ``Display'', ``Door'', ``FoldingChair'', ``Laptop'', ``Microwave'', ``Oven'', ``Phone'', ``Refrigerator'', ``Safe'', ``StorageFurniture'', ``Suitcase'', ``Table'', ``Toilet'', ``TrashCan'', ``WashingMachine'', and ``Window''.
They have significant overlapping with our queries to generate Internet videos, since these objects are common objects which can be articulated.
We control its rotation or translation and render ground truth depth, surface normal, mask, motion type and 3D rotation axis.
The outputs are object articulation videos without backgrounds.

To mimic real 3D scenes, we blend random backgrounds.
Otherwise the detection problem becomes trivial.
We use ScanNet~\cite{dai2017scannet} images with synthetic humans~\cite{varol2017learning} used to train our approach, to ensure the fair comparison.
 
\end{document}